\documentclass[acmsmall]{acmart}
\usepackage{amsmath}
\usepackage{amsthm}
\usepackage{subfig}
\usepackage{hyperref} 
\usepackage{hyperxmp} 
\usepackage{caption}
\usepackage{array}
\usepackage{framed}
\usepackage{graphicx}
\usepackage{color}
\usepackage{multirow}
\usepackage[noend]{algorithmic}
\usepackage{algorithm}

\usepackage[T1]{fontenc}

\usepackage{wrapfig,lipsum,booktabs,graphicx}

\DeclareMathOperator*{\argmin}{arg\,min}

\AtBeginDocument{%
  \providecommand\BibTeX{{%
    \normalfont B\kern-0.5em{\scshape i\kern-0.25em b}\kern-0.8em\TeX}}}

\setcopyright{acmcopyright}
\copyrightyear{2025}
\acmYear{2025}
\acmDOI{XXXXXXX.XXXXXXX}

\acmJournal{CSUR}
\acmVolume{37}
\acmNumber{4}
\acmArticle{111}
\acmMonth{8}

\begin{document}

\title{A Survey on Uncertainty Quantification Methods for Deep Learning}

\author{Wenchong He}
\email{whe.ustc@gmail.com}
\affiliation{%
  \institution{University of Florida}
  \city{Gainesville}
  \state{FL}
  \country{USA}
  \postcode{32611}
}
\author{Zhe Jiang}
\authornote{Contact author: Zhe Jiang, zhe.jiang@ufl.edu}
\email{zhe.jiang@ufl.edu}
\affiliation{%
  \institution{University of Florida}
  \city{Gainesville}
  \state{FL}
  \country{USA}
  \postcode{32611}
}

\author{Tingsong Xiao}
\email{xiaotingsong@ufl.edu}
\affiliation{%
  \institution{University of Florida}
  \city{Gainesville}
  \state{FL}
  \country{USA}
  \postcode{32611}
}

\author{Zelin Xu}
\email{zelin.xu@ufl.edu}
\affiliation{%
  \institution{University of Florida}
  \city{Gainesville}
  \state{FL}
  \country{USA}
  \postcode{32611}
}
\author{Yukun Li}
\email{yukun.li@tufts.edu}
\affiliation{%
  \institution{Tufts University}
  \city{Medford}
  \state{MA}
  \country{USA}
  \postcode{02155}
}




\renewcommand{\shortauthors}{Wenchong He, et al.}

\begin{abstract}
Deep neural networks (DNNs) have achieved tremendous success in computer vision, natural language processing, and scientific and engineering domains. However, DNNs can make unexpected, incorrect, yet overconfident predictions, leading to serious consequences in high-stakes applications such as autonomous driving, medical diagnosis, and disaster response. Uncertainty quantification (UQ) estimates the confidence of DNN predictions in addition to their accuracy. In recent years, many UQ methods have been developed for DNNs. It is valuable to systematically categorize these methods and compare their strengths and limitations. Existing surveys mostly categorize UQ methodologies by neural network architecture or Bayesian formulation, while overlooking the uncertainty sources each method addresses, making it difficult to select an appropriate approach in practice. To fill this gap, this paper presents a taxonomy of UQ methods for DNNs based on uncertainty sources (e.g., data versus model uncertainty). We summarize the advantages and disadvantages of each category, and illustrate how UQ can be applied to machine learning problems (e.g., active learning, out-of-distribution robustness, and deep reinforcement learning). We also identify future research directions, including UQ for large language models (LLMs), AI-driven scientific simulations, and deep neural networks with structured outputs.
\end{abstract}




\maketitle

\section{Introduction}
Deep neural networks (DNNs) have achieved remarkable success in computer vision, natural language processing, and scientific and engineering domains \cite{lecun2015deep,deng2009imagenet}.   Most existing DNNs can be viewed as deterministic functions mapping input features to target predictions through hierarchical representation learning \cite{bengio2013representation}. While DNNs can achieve high accuracy, they sometimes make unexpected, incorrect, yet overconfident predictions, especially in a complex real-world environment \cite{reichstein2019deep}. This has serious consequences in high-stakes applications, such as autonomous driving  \cite{choi2019gaussian}, medical diagnosis  \cite{begoli2019need}, and disaster response \cite{alam2017image4act}. 
In this regard, a DNN model should be aware of what it does not know. 
For example, in the medical domain, when a DNN-based automatic diagnosis system encounters an uncertain case, it should refer the case to a medical expert for more in-depth analysis to avoid fatal mistakes. 
In an autonomous vehicle, if a DNN model knows in what scenarios it tends to make mistakes in estimating road conditions (e.g., bad weather), it can warn the driver to take over and avoid potential crashes.

Recognizing what a DNN model does not know requires assigning appropriate uncertainty scores to its predictions, also called \emph{uncertainty quantification} (UQ).  
Uncertainty in DNNs may come from different types of sources, including data uncertainty and model uncertainty  \cite{yarin2016uncertainty}. Data uncertainty (also aleatoric uncertainty) is an inherent property of the data, which originates from the randomness and stochasticity of the data (e.g., sensor noises) or conflicting evidence between ground truth labels (e.g., class confusion). Data uncertainty is often considered irreducible because we cannot reduce it by adding more training samples. On the other hand, model uncertainty (also epistemic uncertainty) comes from a lack of evidence or knowledge during model training or inference, e.g., limited training samples, sub-optimal DNN model architectures or parameter learning algorithms, and out-of-distribution (OOD) test samples.

\begin{figure*}[ht]
     \centering
     \includegraphics[height=1.4 in]{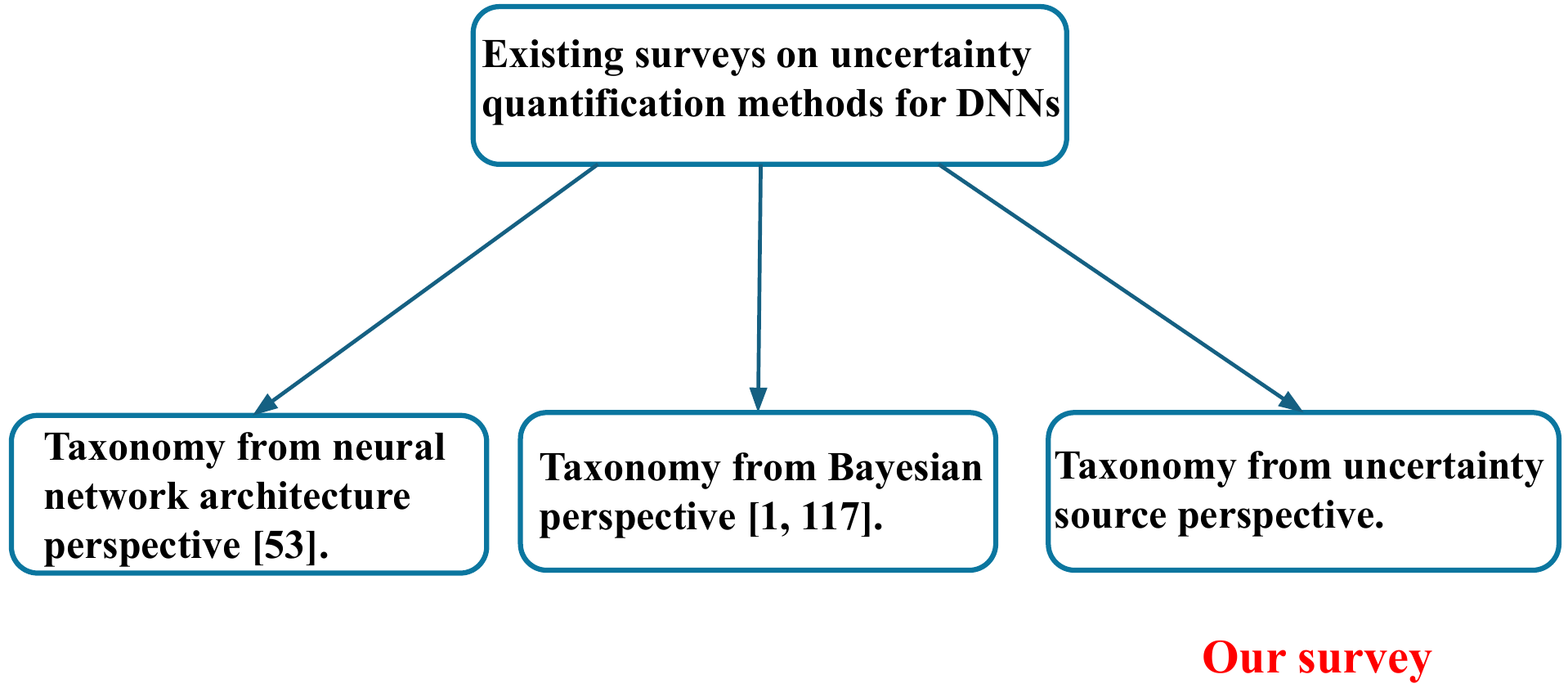}
     \caption{Existing survey on UQ methods for DNNs.}
     \label{fig:survey}
 \end{figure*}

Researchers have recently developed a growing number of UQ methods for DNN models. As shown in Fig.~\ref{fig:survey}, existing surveys of UQ methods for DNNs typically categorize approaches from either the neural network architecture perspective or the Bayesian modeling perspective.  Specifically, Gawlikowski et al. {\cite{gawlikowski2021survey}} categorize existing UQ methods based on their types of DNN model architectures, including Bayesian neural networks, ensemble models, and single model architectures, without discussing the connection between DNN model architectures and the types of uncertainty they address. Other surveys only focus on the Bayesian perspective. For example,  Mena et al. {\cite{mena2021survey}} provide a comprehensive review of Bayesian neural networks for UQ but overlook methods from a frequentist perspective (e.g., prediction interval, ensemble methods). Abdar et al. { \cite{abdar2021review}} cover the ensemble methods and other frequentist methods, but the paper does not compare their advantages and disadvantages.  To the best of our knowledge, existing surveys on UQ methods often overlook the types of uncertainty sources these methods address. This perspective is important for selecting the appropriate UQ methods for different applications where one type of uncertainty source dominates others. 

To fill the gap, we provide the first survey with a systematic taxonomy of UQ methods for DNNs from the perspective of different uncertainty sources, including model uncertainty, data uncertainty, and both. We summarize the characteristics of different methods and compare their advantages and disadvantages. We also connect the taxonomy to several major deep learning topics where UQ is crucial. Finally, we identify research gaps and suggest several future research directions. 
The overall structure of this survey  is as follows:
\begin{itemize}
    \item Section 2 defines two different types of uncertainty sources, i.e., data uncertainty and model uncertainty, in the supervised learning setting. This lays a foundation for our discussion on various UQ methods.
    \item Section 3 highlights the practical applications of uncertainty quantification (UQ) for deep learning, focusing on how UQ applies to various real-world problems, such as medical diagnosis, geosciences, and transportation. 
    \item Section 4 presents a taxonomy of UQ methods for deep learning based on the types of uncertainty sources they capture, including data uncertainty, model uncertainty, and both. 
    \item Section 5 discusses how UQ plays a significant role in several key machine learning problems (OOD detection, active learning, deep reinforcement learning). 
    \item Section 6 discusses several future research directions, including UQ for large language models (LLMs), UQ for DNNs in scientific simulations, UQ for DNNs with structured outputs (e.g., spatiotemporal data and graphs), and combining UQ with explainability.
\end{itemize}

\section{Types of uncertainty source}\label{typeuq}


This section starts with the mathematical formulation of supervised learning. Based on that, we define two sources of uncertainty: data uncertainty and model uncertainty.

{\color{black} \subsection{Preliminaries of Supervised Learning}\label{sec:sup}

Given training data $\mathcal{D}_{tr} = \{(\boldsymbol{x}_i,{y}_i)\}_{i=1}^{n} \subset \mathcal{X}\times \mathcal{Y}$,
$\mathcal{X}\subseteq \mathbb{R}^d$ is the input sample feature space, and $\mathcal{Y}$ is the target variable space, where $\mathcal{Y}=\{\omega_1,...,\omega_k\}$ for a classification problem with $k$ classes, and $\mathcal{Y} \subseteq \mathbb{R}$ for a regression problem. Each training sample is assumed to be independent and identically distributed (i.i.d.) from some unknown probability distribution $p(\boldsymbol{x}, y)$ on the space $\mathcal{X}\times\mathcal{Y}$. Given a hypothesis space $\mathcal{H}$ consisting of hypotheses $h: \mathcal{X} \rightarrow \mathcal{Y}$ and a loss function $l$ that measures the discrepancy between prediction and ground-truth, a learning problem aims to find the best hypothesis in the hypothesis space that minimizes the loss \cite{hullermeier2021aleatoric}:

\begin{equation} \label{eq:lossfun} \footnotesize
{h^\ast} = \argmin_{h \in \mathcal{H}} R(h), \quad \text{where} \quad
R(h) = \int {l}(y, h(\boldsymbol{x})) p(\boldsymbol{x}, y) d\boldsymbol{x}dy.
\end{equation}

In practice, the model is learned by minimizing the empirical risk \cite{murphy2012machine}, defined as the average loss over the training data $\mathcal{D}_{tr}$:

\begin{equation}\label{eq:emploss}\footnotesize
\tilde{h} =   \argmin_{h\in \mathcal{H}}  R_{emp}(h), \  \text{where} \ 
  R_{emp}(h) = \frac{1}{|\mathcal{D}_{tr}|}\sum_{(\boldsymbol{x}_i,y_i)\in\mathcal{D}_{tr}}{l}(y_i, h(\boldsymbol{x}_i)).
\end{equation}


\subsection{Model uncertainty}

\subsubsection{Sources of model uncertainty}
{\color{black} Model uncertainty (a.k.a. epistemic uncertainty) represents the uncertainty in a model's predictions related to the imperfect model training process. It is reducible given more training data.  There are several common types of model uncertainty: {uncertainty in the choice of model family, uncertainty in model parameter learning, and uncertainty due to different sample distributions between model training and model inference  (e.g., out-of-distribution samples).   These types of model uncertainty are illustrated in Fig.~\ref{fig:modeluq},  where  $\mathcal{F}$ denotes the entire \begin{wrapfigure}{r}{5.5cm} 
    \centering
{\includegraphics[height=1.2in]{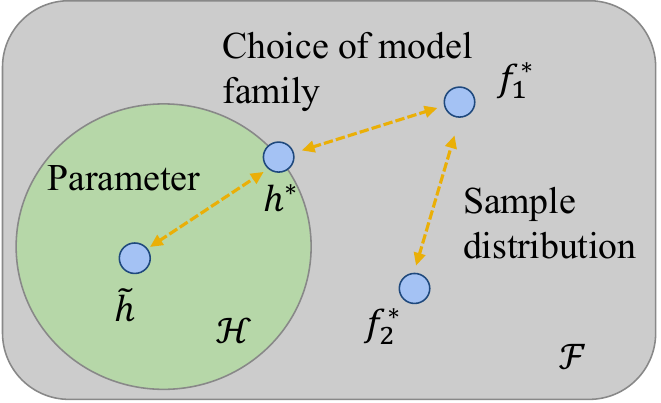}} 
    \caption{\color{black} Visualization on various model uncertainty sources.}
    \label{fig:modeluq}
    \vspace{-5mm}
\end{wrapfigure}  hypothesis space, $f_1^*$ and $f_2^*$ are the theoretical optimal hypotheses  within $\mathcal{F}$ (based on Eq.~\ref{eq:lossfun}) for two different sample distributions $p_1(\boldsymbol{x}, y)$ and $p_2(\boldsymbol{x}, y)$, respectively. That is, ${f}_i = \argmin_{h \in \mathcal{F}} \int {l}(y, h(\boldsymbol{x})) p_i(\boldsymbol{x}, y) d\boldsymbol{x}dy \ \text{for} \  i=1, 2$.
$\mathcal{H}$ is the sub-hypothesis space for one particular model architecture and set of hyperparameters (e.g., a specific transformer architecture). $h^*$ is the theoretical optimal solution within $\mathcal{H}$ for a sample distribution $p_1(\boldsymbol{x}, y)$ based on Eq.~\ref{eq:lossfun}. $\tilde{h}$ is the empirical solution within $\mathcal{H}$ that is learned by an optimizer based on a particular training data $\mathcal{D}_{tr}$ drawn from the population distribution $p_1(\boldsymbol{x}, y)$ (see Eq.~\ref{eq:emploss}).

Table~\ref{table:modeluqsource} summarizes the different sources of model uncertainty in the supervised learning framework. The first type, the choice of model family, is due to the lack of knowledge of which type of model architecture is the most suitable. Because of this, the theoretical optimal solution  
$h^*$ within $\mathcal{H}$  (assuming a particular model architecture) is different from the theoretical optimal $f_1^*$. It is related to the ``bias" part in the bias-variance decomposition \cite{gruber2023sources}. 
 Second, due to limited training data or imperfect parameter learning algorithms, the learned model $\tilde{h}$ based on the empirical loss may exhibit 
 \begin{wraptable}{r}{0.50\textwidth} 
\footnotesize
    \centering
    \small
    \captionof{table}{\color{black} Comparison of different types of model uncertainty in the supervised learning setting}
    \begin{tabular}{|c|c|}
\hline
\textbf{\begin{tabular}[c]{@{}c@{}}Model uncertainty\\  sources\end{tabular}} & \textbf{\begin{tabular}[c]{@{}c@{}}Corresponding notation\\  in supervised learning\end{tabular}} \\ \hline
\begin{tabular}[c]{@{}c@{}}{Choice of model} \\ {family}\end{tabular}                & \begin{tabular}[c]{@{}c@{}}Optimal solution $h^*$ within \\ $\mathcal{H}$ does not align \\ with theoretical \\ optimal $f^*$ in $\mathcal{F}$\end{tabular}                                   \\ \hline
\begin{tabular}[c]{@{}c@{}}Model \\ parameter learning\end{tabular}                  & \begin{tabular}[c]{@{}c@{}} Learned solution $\tilde{h}$ \\ does not align with \\ optimal $h^*$ in $\mathcal{H}$ \end{tabular}    \\ \hline
\begin{tabular}[c]{@{}c@{}} Different sample \\ distributions in \\learning and inference\end{tabular}           & \begin{tabular}[c]{@{}c@{}} Theoretical optimum \\ $f_1^*$ and $f_2^*$ mismatch\\ under different sample\\  distribution $p(\boldsymbol{x},y)$\end{tabular}                                             \\ \hline
\end{tabular}
      
      \label{table:modeluqsource}
\end{wraptable}
variations and deviate from the theoretically optimal solution  $h^*$ within $\mathcal{H}$.  This leads to model uncertainty related to model parameter learning. 
It is related to the ``variance" part in the bias-variance decomposition. 
Third, model uncertainty can be related to the different sample distributions between model training and inference. For example, there may be two different sample distributions $p_1(\boldsymbol{x}, y)$ and $p_2(\boldsymbol{x}, y)$. Their theoretical optimal solutions $f_1^*$ and $f_2^*$ are different. Because of this, a model learned from training samples following $p_1(\boldsymbol{x}, y)$ will contain uncertainty in its inference on a sample drawn from   $p_2(\boldsymbol{x}, y)$, i.e., out-of-distribution (OOD) samples. Another relevant scenario is that when a model makes predictions on a test sample that is far away from other training samples (or surrounded by sparse training samples) in the feature space $\mathcal{X}$ (see Fig.~\ref{fig:gpvis}), the prediction tends to have higher uncertainty. This scenario is also relevant to the OOD case since a test sample that is far away from other training samples is more likely to be an OOD sample. Note that our definition of the three types of model uncertainty may overlap. For instance, model uncertainty due to the lack of training samples near a test sample can also be considered model parameter learning uncertainty. 

}

\subsubsection{Model uncertainty representation}
{\color{black} 
As discussed above, the sources of model uncertainty can arise from different aspects: the choice of model hyper-parameters, model parameter learning, and different sample distributions in learning and inference. In general, there are various ways to represent model uncertainty from each type. First, uncertainty in model parameter learning arises from suboptimal parameter optimization. To account for this uncertainty source, one approach is 
through a Bayesian neural network (BNN) \cite{jospin2022hands}.  A BNN assumes a prior over the model parameters and aims to infer the posterior distribution of the model parameters to reflect the parameter uncertainty. This provides a theoretical foundation for model uncertainty. Second, uncertainty arising from the choice of model hyper-parameters is due to the inductive bias in choosing a sub-hypothesis space (e.g., a particular DNN architecture) \cite{zaidi2021neural,wenzel2020hyperparameter,yang2022calibrate}. It can be estimated with ensembles of different neural network architectures (deep ensembles) \cite{lakshminarayanan2017simple}. The intuition is to construct an ensemble of neural network architectures, each of which is trained separately. The predictions of the ensemble on an input form a distribution over the target variable. Thus, the variance of the target variable predictions can be used to estimate the prediction uncertainty.  
The third type of uncertainty arises from differences in sample distributions, which are caused by the mismatch between the distribution of the training dataset and that of a test sample. Capturing this type of uncertainty requires learning meaningful embeddings that reflect sample distances.  More details are discussed in Section~\ref{sec:texonomy}.}

}

 \subsection{Data uncertainty}
\subsubsection{Source of data uncertainty}

\begin{figure}
    \centering
   \subfloat[ {\color{black} Sample distribution with  separable class boundary.}]{ \includegraphics[height=0.9in]{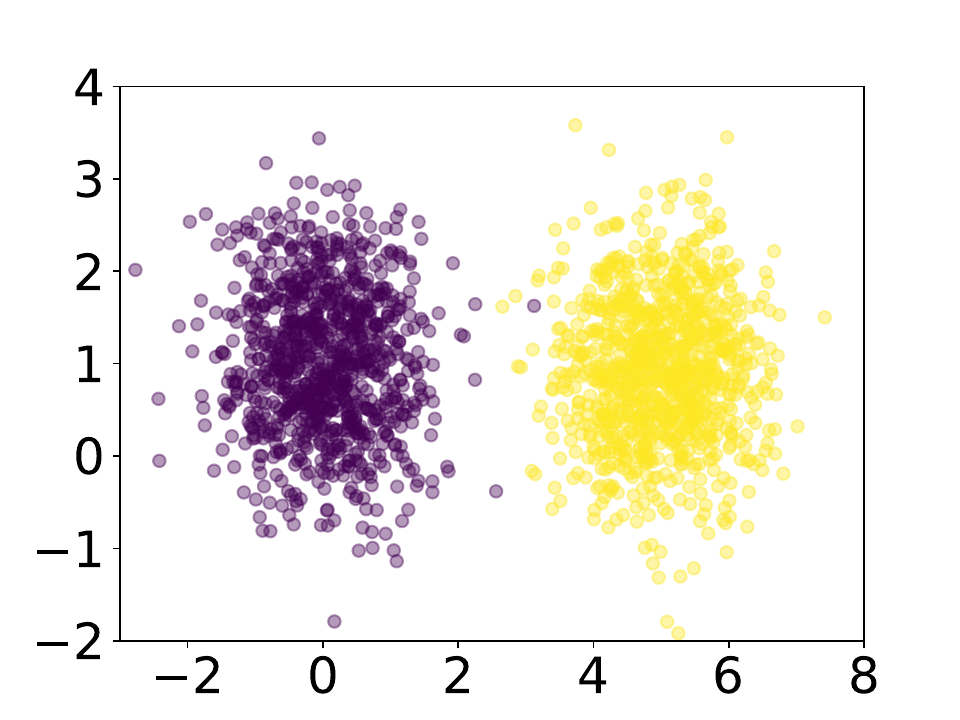}} \hspace{0.1in}
   \subfloat[{\color{black} Entropy of sample distribution in (a).}]{\includegraphics[height=0.9in]{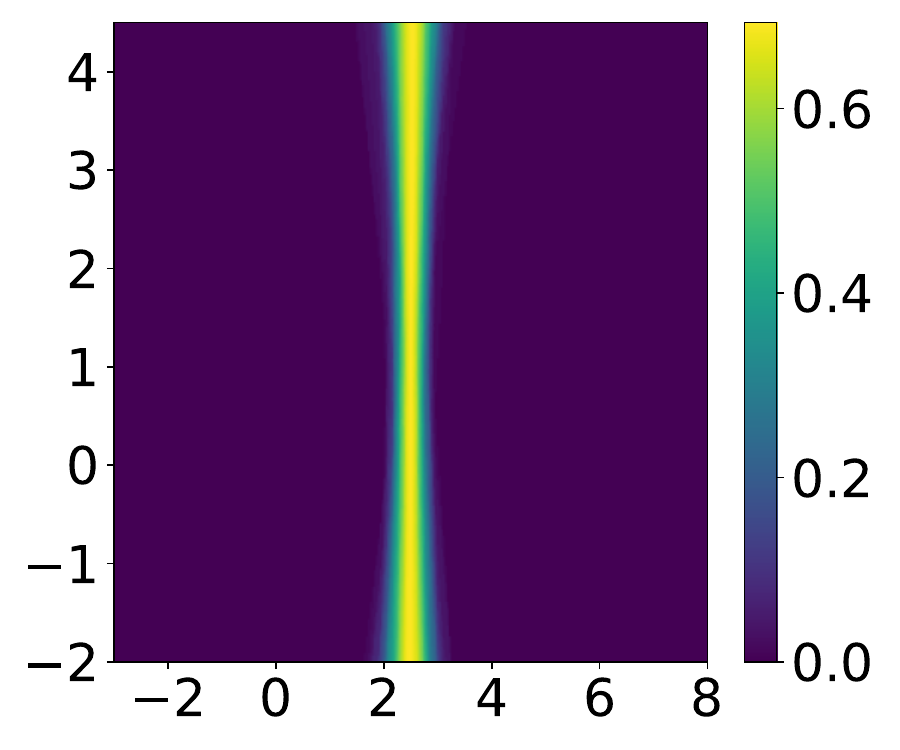}}  
   \hspace{0.1in}
   \subfloat[{\color{black} Sample distribution with ambiguous class boundary.}]{\includegraphics[height=0.9in]{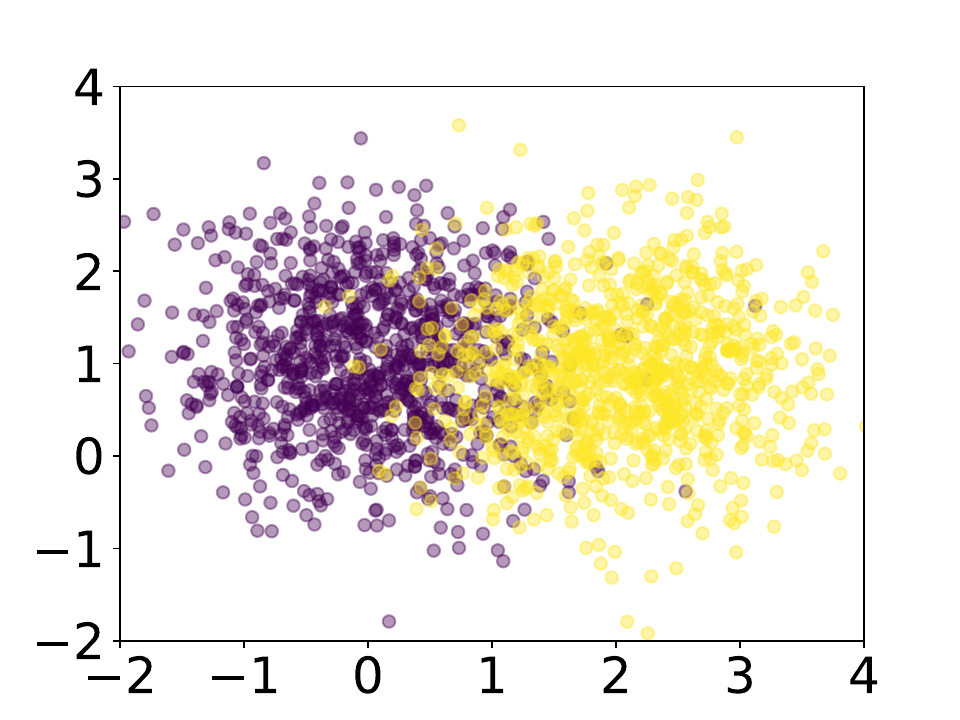}} \hspace{0.1in}
   \subfloat[{\color{black}Entropy of sample distribution in (c).}]{\includegraphics[height=0.9in]{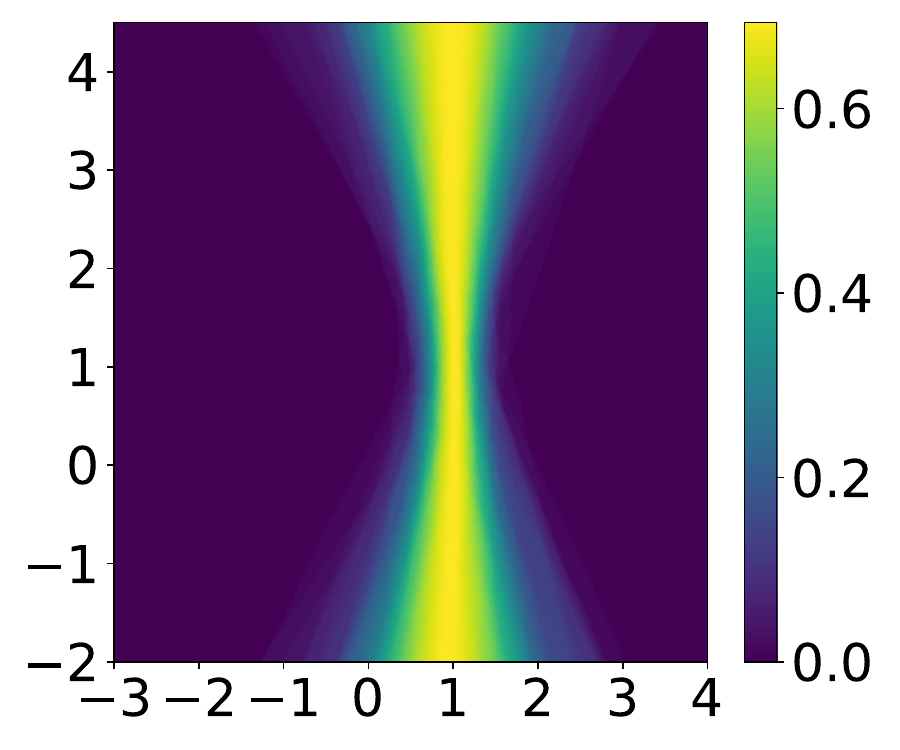}}
    \caption{Data uncertainty visualization examples (Different colors represent samples in different classes).}
    \label{fig:entropy_uq}
\end{figure}

{\color{black}

Data uncertainty (a.k.a aleatoric uncertainty) arises from inherent data randomness, noise, or class confusion (i.e., the same feature value can correspond to different classes in the sample distribution). It is irreducible even with more training data \cite{yarin2016uncertainty}.  Randomness or noise in data can arise in data acquisition due to instrument errors, data transmission errors, and inappropriate data storage and formatting \cite{hariri2019uncertainty}.}
For example, for spatiotemporal data collected from various space and airborne platforms (e.g., CubeSat, UAVs), the data uncertainty may result 
from the sensor errors associated with the data acquisition devices and the fact that data are acquired in a digital format (which is discrete in nature) \cite{cheng2014managing}
 even though the underlying process is continuous.

{\color{black}
\subsubsection{Data uncertainty representation}
Consider a training dataset $\mathcal{D}_{tr}$ drawn from the distribution ${p(\boldsymbol{x},y)}$. Several techniques exist for representing data uncertainty to account for the inherent randomness in the mapping from $\boldsymbol{x}$ to $y$. In the context of a discriminative classification task, one method to represent the uncertainty of the class variable, given a specific input $\boldsymbol{x}$ is the maximum class probability $\max_y p(y|\boldsymbol{x})$. Another approach uses the entropy of the condition class distribution $p(y|\boldsymbol{x})$, which captures the randomness of the class distribution due to class confusion. For high-dimensional structured samples, deep generative models can be employed to learn the  complex underlying  distribution of the data and quantify uncertainty. 



Data uncertainty arises from natural variability in data (for regression) and class confusion (for classification). Consider the toy distribution in Fig.~\ref{fig:entropy_uq} as an example for classification, which consists of two normally distributed clusters. Each cluster (color) represents a separate class. The dataset in Fig.~\ref{fig:entropy_uq} (a) has a sharper class boundary, indicating lower data uncertainty. The entropy of most samples is low except for those near the class boundary, as shown in Fig.~\ref{fig:entropy_uq} (b).  In contrast, the dataset in Fig.~\ref{fig:entropy_uq} (c) exhibits more confusion between the two classes, corresponding to higher data uncertainty  as shown in Fig.~\ref{fig:entropy_uq} (d).

Beyond class confusion, data uncertainty can also arise from  inherent noise (variability) in the data generation or collection process. For example, in a regression problem, the observations can be represented by: $y = f(\boldsymbol{x}) + \epsilon (\boldsymbol{x})$, where $f(\boldsymbol{x})$ is the true data generation function, and
  $\epsilon (\boldsymbol{x})$ represents the measurement noise. There are two classes of noise: homoscedastic and heteroscedastic noise \cite{kendall2017uncertainties}. Homoscedastic noise assumes constant noise variance across all the \begin{wrapfigure}{r}{5.7cm} 
    \centering
    \includegraphics[height=1.4in]{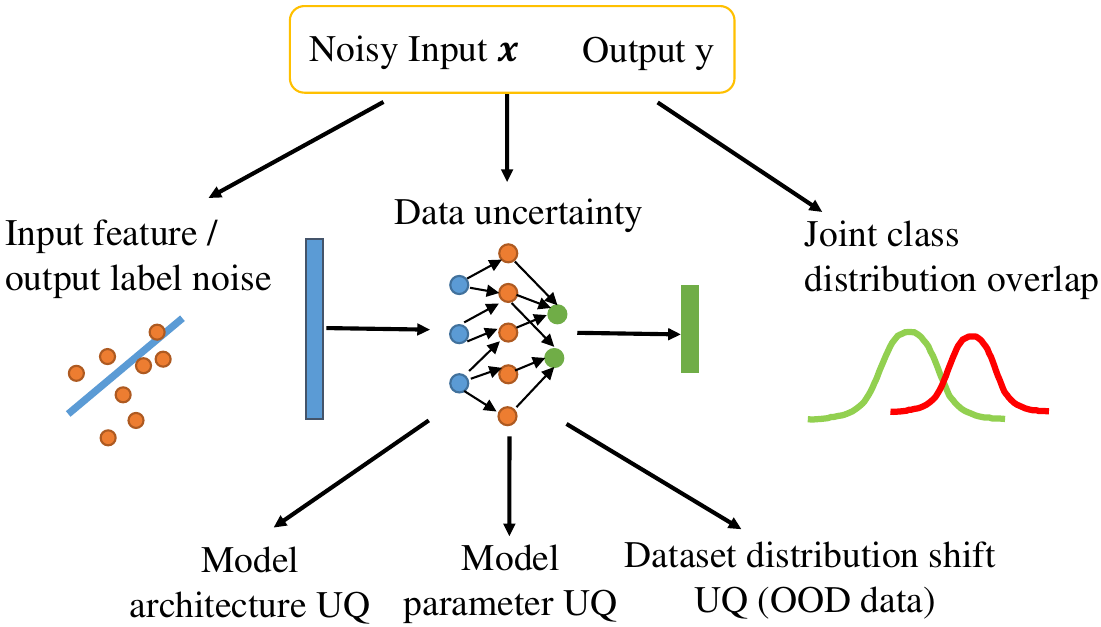}
    \caption{Different types of uncertainty source.}
    \label{fig:uqsource}
\end{wrapfigure}
$\boldsymbol{x}$ inputs. Heteroscedastic noise, on the other hand, models the observation noise as a function of the input $\epsilon(\boldsymbol{x}) \sim p(\epsilon|\boldsymbol{x})$ (e.g., heteroscedastic Gaussian noise). The heteroscedastic noise model is useful in the case where the noise level varies for different samples. 
  
In summary, we have described the sources and representations of both model and data uncertainty. As Fig.~\ref{fig:uqsource} shows, data uncertainty arises from the inherent properties of the given data, while model uncertainty stems from issues such as misspecification of model architectures, parameters, and the differences in sample distributions. Depending on the nature of the application, the predominant source of uncertainty may vary.




}
\section{Application domains}

In this section, we discuss several application domains of uncertainty quantification for deep learning models. For each application, we discuss the motivation for developing uncertainty-aware models, the source of uncertainty, and the challenges associated with uncertainty quantification. The applications in medical diagnosis, geoscience, transportation and natural language processing are discussed below. Additional applications in Biochemistry engineering and engineering design are discussed in appendix. 

\textit{Medical diagnosis}: 
DNN models have achieved tremendous success in various medical applications, including medical imaging, clinical diagnosis support, and treatment planning \cite{oh2020deep}. However, a critical concern is that deep learning models tend to be over-confident even for a wrong prediction \cite{loftus2022uncertainty}, which can lead to serious consequences. Thus, it is essential to estimate the prediction uncertainty (confidence). 
Both data uncertainty and model uncertainty exist in medical problems. Data uncertainty arises from noisy measurements from medical devices, ambiguous class labels (e.g., non-consensus tumor boundary annotations between different radiologists), and registration errors between medical imagery taken at different times or from different devices~\cite{gong2022uncertainty}. 
Model uncertainty also exists because patient cases in the test cases may not be well-represented in the training set. 
There are several challenges in developing UQ methods. First, medical data contains diverse sources of noise and uncertainty. Second, 
Interpretability in uncertainty quantification is important, but it remains an unsolved issue in medical problems.
Existing uncertainty-aware deep learning models in medical domains can be categorized into those related to medical imaging and those for non-medical imaging applications \cite{loftus2022uncertainty}. In medical imaging, deep learning is often used for segmentation or classification of magnetic resonance imaging, ultrasound, and coherence tomography  imagery~\cite{edupuganti2020uncertainty}. These studies often focus on data uncertainty due to ambiguous labels \cite{qin2021super}, or image registration uncertainty \cite{chen2021deep}. 
Non-medical imaging applications are mostly related to clinical diagnosis support and treatment planning from Electronic Health Records. 
The presence of significant variability in personalized predictions \cite{dusenberry2020analyzing} requires a model to capture prediction uncertainty.

\textit{Geoscience}: With advances in GPS and remote sensing technologies, a growing volume of spatiotemporal data is being collected from spaceborne, airborne, seaborne, and terrestrial platforms~\cite{shekhar2015spatiotemporal}. Emerging spatiotemporal big data, increased computational power (GPUs), and  recent advances in deep learning technologies provide unique opportunities to advance our knowledge of the Earth system~\cite{shekhar2015spatiotemporal}. For example, deep learning has been used to predict river flow and temperature~\cite{jia2021physics} and hurricane tracks~\cite{kim2019deep}. Uncertainty quantification for deep learning is important in geoscience because of the high-stakes decision-making involved (e.g., evacuation planning based on hurricane tracking with a “cone of uncertainty”).  
Several challenges arise from the unique characteristics of spatiotemporal data. First, spatiotemporal data exhibit various spatial, temporal, and spectral resolutions and diverse sources of noise and errors (e.g., sensor noise, obstacles, atmospheric effects in remote sensing signals~\cite{licata2022uncertainty}, and GPS errors). 
Second, spatial registration errors and uncertainties may arise when co-registering different layers of geospatial data into the same spatial reference system \cite{he2022quantifying}. Third, spatiotemporal data are heterogeneous, i.e., the data distribution often varies across different regions or time periods \cite{jiang2019spatial}. As a result, a deep learning model trained in one region or time period may not generalize well to another. This issue is particularly significant when spatial observations are sparsely distributed, leading to uncertainty in inferring values at other locations in continuous space.


\textit{Transportation}:
Deep learning applied to transportation data (e.g., ground sensors and video cameras on the road) provides unique opportunities to monitor traffic conditions, analyze traffic patterns, and improve decision-making.  For instance, temporal graph neural networks are used to predict traffic flows (e.g., congestion or accidents), and incorporating physical principles into neural network modeling further enhances  traffic modeling performance \cite{ ji2022stden}.   Deep learning plays a critical role in autonomous driving (e.g., using LiDAR sensors and optical cameras to detect road lanes, other vehicles, or pedestrians). Uncertainty quantification for AI in transportation is challenging due to  temporal dynamics, the complexity of road environment, and the existence of noise and uncertainty (e.g., omission, sparse sensor coverage, errors, or inherent biases).  For example, highly crowded events can disrupt normal traffic flows on road networks.  
 Existing studies on trajectory prediction uncertainty consider the data uncertainty due to  sparse or
insufficient training data, and erroneous or missing
measurements from signal loss \cite{markos2021capturing}.  Other studies consider complex environmental factors such as extreme weather conditions \cite{wang2019uncertainty}. Uncertainty in short-term traffic status forecasting (e.g., volume, travel speeds, and occupancy)  is related to the stochastic environment and model training \cite{wang2014new}, while uncertainty in long-term traffic modeling stems from  exogenous factors affecting traffic flow (e.g., rainstorms and snowstorms) \cite{li2022quantifying}.

{\color{black}  

\textit{Natural Language Processing:} The advancement of pre-trained language models (PLMs) has revolutionized language processing by addressing various tasks in a unified manner (e.g., machine translation, sentiment analysis, speech recognition)~\cite{li2021pre}. Although natural language processing (NLP) has made remarkable strides with the emergence of large language models (LLMs), these models are prone to hallucinations (i.e., generating misleading or fabricated content) \cite{xu2024hallucination}. Thus, uncertainty quantification plays a critical role in improving the trustworthiness of LLMs. However, quantifying uncertainty in LLMs presents significant challenges. First, uncertainty can arise from various sources related to both data and models. Data uncertainty stems from ambiguities, noise (e.g., out-of-vocabulary words or distractions), and semantic complexities in the input language. Model uncertainty occurs when the model lacks the specific knowledge required for out-of-distribution input queries, leading to arbitrary responses. Second, the large vocabulary space complicates the direct assessment of confidence through probability likelihood, and relying solely on prediction logits can lead to overconfidence \cite{guo2017calibration}. Therefore, it is essential to capture uncertainty in the semantic space due to the high semantic similarity among different possible answers. Third, aligning uncertainty with actual correctness (calibration) is crucial for trustworthy applications of LLMs.



}


\section{A Taxonomy of UQ Methodologies for DNNs}\label{sec:texonomy}

In this section, we provide a new taxonomy (Fig.~\ref{fig:taxonomy}) of UQ methods for DNNs based on the type of uncertainty sources: model uncertainty, data uncertainty, and the combination of the two. We discuss the underlying intuitions of methods in each category and compare their pros and cons. 




\begin{figure*}[ht]
    \centering
    \includegraphics[height=1.4in]{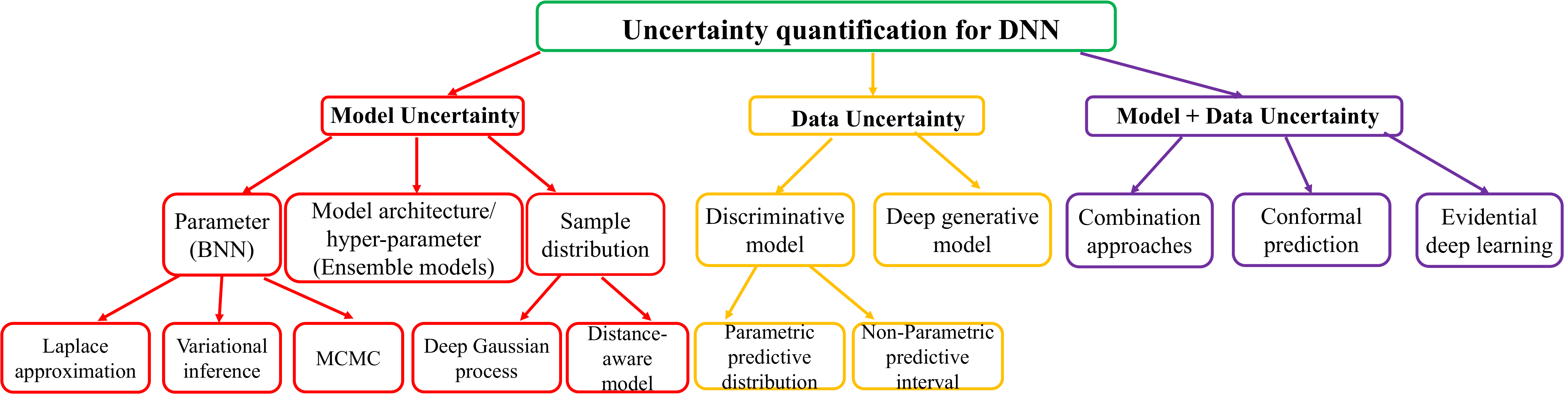}
    \caption{\color{black} A taxonomy for existing literature on UQ for DNN.}
    \label{fig:taxonomy}
\end{figure*}


\subsection{\bf Model Uncertainty }



This subsection reviews the existing methods for model uncertainty in DNNs. We categorize these methods into three subcategories: Bayesian neural network, ensemble models, and sample distribution-related models. We now introduce each subcategory.

\subsubsection{Bayesian Neural Networks}
From a frequentist point of view, there exists a single set of parameters $\boldsymbol{\boldsymbol{\theta}}^*$ that best fit the DNN model, where $\boldsymbol{\theta}^* = \argmin_{\boldsymbol{\theta}}\mathcal{L}(\mathbf{Y}, f(\mathbf{X}, \boldsymbol{\theta}))$ and $\mathcal{L}$ is the loss function. However, the point estimation of DNN parameters can be over-fitting and overconfident \cite{thulasidasan2019mixup}. 
In contrast, the Bayesian neural network (BNN) imposes a prior on the neural network parameters $p(\boldsymbol{\theta})$ and learns the posterior distribution of these parameters $p(\boldsymbol{\theta}|\mathbf{X}, \mathbf{Y})= \frac{p(\mathbf{Y}|\mathbf{X},\boldsymbol{\theta} )p(\boldsymbol{\theta})}{p(\mathbf{Y}|\mathbf{X})}$. 
This term represents the posterior distribution of model parameters conditioned on the training dataset. This distribution reflects the extent to which our model can capture patterns in the training data. Assuming a Gaussian distribution for the model parameters, a larger variance in the distribution indicates greater uncertainty in the model. Such uncertainty can be caused by a limited amount of training data. For inference on  new samples $\boldsymbol{x}^*$, we can marginalize out the model parameters as follows:
\begin{equation}\label{eq:bayesian}\footnotesize
    p(y^*|\boldsymbol{x}^*,\mathbf{X}, \mathbf{Y} ) = \int p(y^*|\boldsymbol{x}^*, \boldsymbol{\theta})p(\boldsymbol{\theta}|\mathbf{X}, \mathbf{Y}) d\boldsymbol{\theta}.
\end{equation}
The uncertainty of a test sample is reflected by the variance of its prediction distribution.  {\color{black} Although BNN can theoretically quantify total uncertainty by modeling both prediction distribution $p(y^*|\boldsymbol{x}^*, \boldsymbol{\theta})$ (data uncertainty)  and parameter posterior distribution $p(\boldsymbol{\theta}|\mathbf{X}, \mathbf{Y})$ (model uncertainty), most existing work simplifies the analysis by treating the prediction $p(y^*|\boldsymbol{x}^*, \boldsymbol{\theta})$ as deterministic. Therefore, these BNN methods primarily capture parameter uncertainty rather than total uncertainty.}

However, the parameter posterior distribution is analytically intractable and lacks a closed-form solution. An approximation must be made. Various approaches have been proposed to estimate the posterior of the neural network parameters in a simpler and tractable form.
Some approximation methods define a parameterized class of distributions, $\mathcal{Q}$, from which they select an approximation $q_{\phi}(\boldsymbol{\theta})$ for the posterior. For example, $\mathcal{Q}$ can be the set of all factorized Gaussian distributions, and $\phi$ is the parameters of the mean and diagonal variance. The distribution $q_{\phi}(\boldsymbol{\theta}) \in \mathcal{Q}$ is selected according to some optimization criteria to approximate the posterior. Two popular methods for optimization are \textit{variational inference} \cite{blei2017variational} and \textit{Laplace approximation} \cite{friston2007variational}. Instead of approximating the posterior analytically, another approach is to address this problem using Monte Carlo sampling, specifically \textit{Markov Chain Monte Carlo sampling}. 

 

{\bf Variational Inference (VI)}:
     In~ \cite{hinton1993keeping,graves2011practical}, the authors propose finding a variational approximation to the
Bayesian posterior distribution on the weights by maximizing the \textit{evidence lower bound} (ELBO)  of the log marginal likelihood. The intractable posterior $p(\boldsymbol{\theta}|\mathbf{X}, \mathbf{Y})$ is approximated with a parametric  distribution $q_{\boldsymbol{\phi}}(\boldsymbol{\theta})$. 

\begin{equation}\label{eq:vi}\footnotesize
    \begin{split}
        \log p(\mathbf{Y}|\mathbf{X}) &\geq \mathbb{E}_{\boldsymbol{\theta} \sim q_{\phi}(\boldsymbol{\theta})} \log \frac{p(\mathbf{Y}|\mathbf{X}, \boldsymbol{\theta})p(\boldsymbol{\theta})}{q_{\phi}(\boldsymbol{\theta})} 
        = \mathbb{E}_{\boldsymbol{\theta} \sim q_{\phi}(\boldsymbol{\theta})}\log p(\mathbf{Y}|\mathbf{X}, \boldsymbol{\theta}) - \mathcal{KL}(q_{\phi}(\boldsymbol{\theta})||p(\boldsymbol{\theta})).
    \end{split}
\end{equation}

where the first term $\mathbb{E}_{\boldsymbol{\theta} \sim q_{\phi}(\boldsymbol{\theta})}\log p(\mathbf{Y}|\mathbf{X}, \boldsymbol{\theta}))$ is log-likelihood of the training data on the neural network model. This term increases as the model accuracy increases. The second term is the \textit{Kullback-Leibler} (KL) divergence between the posterior estimation and the prior of the model parameters, which controls the complexity of the neural network model. Maximizing Eq.~\ref{eq:vi} corresponds to finding a tradeoff between the prediction accuracy and complexity of the model. Thus the posterior inference problem becomes an optimization problem on the parameter $\boldsymbol{\phi}$.

One challenge of variational inference lies in the choice of the parameterized class of distribution $q_{\boldsymbol{\phi}}(\boldsymbol{\theta})$. The original method relies on the use of a Gaussian approximating distribution with a \textit{diagonal} covariance matrix (mean-field variational inference)  \cite{hinton1993keeping,posch2019variational}. This method leads to a straightforward lower bound for optimization, but the approximation capability is limited.  To capture the posterior correlations between parameters, the second category of approaches extends the diagonal covariance to a \textit{general} covariance matrix while still leading to a tractable algorithm by maximizing the above ELBO \cite{posch2020correlated}. However, the full covariance matrix not only increases the number of trainable parameters but also introduces substantial memory and computational cost for DNN. To reduce the computation, the third category of approaches aims to simplify the covariance matrix structure with certain assumptions. Some approaches assume independence among layers, resulting in \textit{block-diagonal structure} covariance matrix \cite{sun2017learning,zhang2018noisy}.
Others find that for a variety of deep BNN trained using Gaussian variational inference,  the posterior consistently exhibits strong low-rank structure after convergence  \cite{swiatkowski2020k}, and they propose to decompose the dense covariance matrix into a low-rank factorization to simplify the computation. 
Additionally,  \cite{mishkin2018slang} assumes the covariance matrix takes the form of \textit{“diagonal plus low-rank"}  structure for a more flexible and faster approximation.
Another line of work introduces sparse uncertainty structure via a hierarchical posterior or employing normalizing flow with low-dimensional auxiliary variables \cite{ritter2021sparse,louizos2017multiplicative} to reduce the computation.

The disadvantage of variational Gaussian approximation for DNN parameters is that to capture the full correlations among model latent weights, it requires a large number of variational parameters to be optimized, which scales quadratically with the
number of latent weights in the model. Existing works aim to simplify the covariance structure with certain assumptions while still capturing the correlation between neural network parameters to optimize the computational speed.

 {\bf Laplace approximation:} The idea behind the Laplace approximation is to obtain an approximate posterior around the \textit{'Maximum A Posterior'} (MAP) estimator of neural network weights with a Gaussian distribution, based on the second derivative of the neural network likelihood functions \cite{mackay1992practical}. The method can be applied post hoc to a pre-trained neural network model. 
\begin{equation}\footnotesize
    p(\boldsymbol{\theta}|\mathbf{X},\mathbf{Y}) \approx p(\boldsymbol{\hat{\theta}}|\mathbf{X},\mathbf{Y})\exp(-\frac{1}{2}(\boldsymbol{\theta} - \boldsymbol{\hat{\theta}})^T\mathbf{H}(\boldsymbol{\theta} - \boldsymbol{\hat{\theta}})).
\end{equation}

Thus, the posterior is then approximated as a Gaussian:
\begin{equation}\footnotesize
    p(\boldsymbol{\theta}|\mathbf{X},\mathbf{Y}) = \mathcal{N}(\boldsymbol{\hat{\theta}}, \mathbf{H}^{-1}).
\end{equation}

Where $\boldsymbol{\hat{\theta}}$ is the MAP estimate and  $\mathbf{H}$ is the Hessian matrix. It is the second derivative of the neural network likelihood function with respect to the model weights, i.e., $\mathbf{H}_{ij}= -\frac{\partial^2}{\partial \boldsymbol{\theta}_i \partial \boldsymbol{\theta}_j}\log p(\mathbf{Y}|\mathbf{X}, \boldsymbol{\theta})$. Fig. ~\ref{fig:laplace} (a) illustrates the intuition of Laplace approximation, where the blue density function is the true posterior distribution and the orange Gaussian distribution is the Laplace approximation. This method approximates the posterior locally and can simplify the computation of the posterior, but the downside is that it cannot capture the multi-modal distribution with more than one mode since it is a local estimation around the MAP mode. 



\begin{figure}[h]
    \centering
    \subfloat[Laplace approximation]{\includegraphics[height = 0.8in]{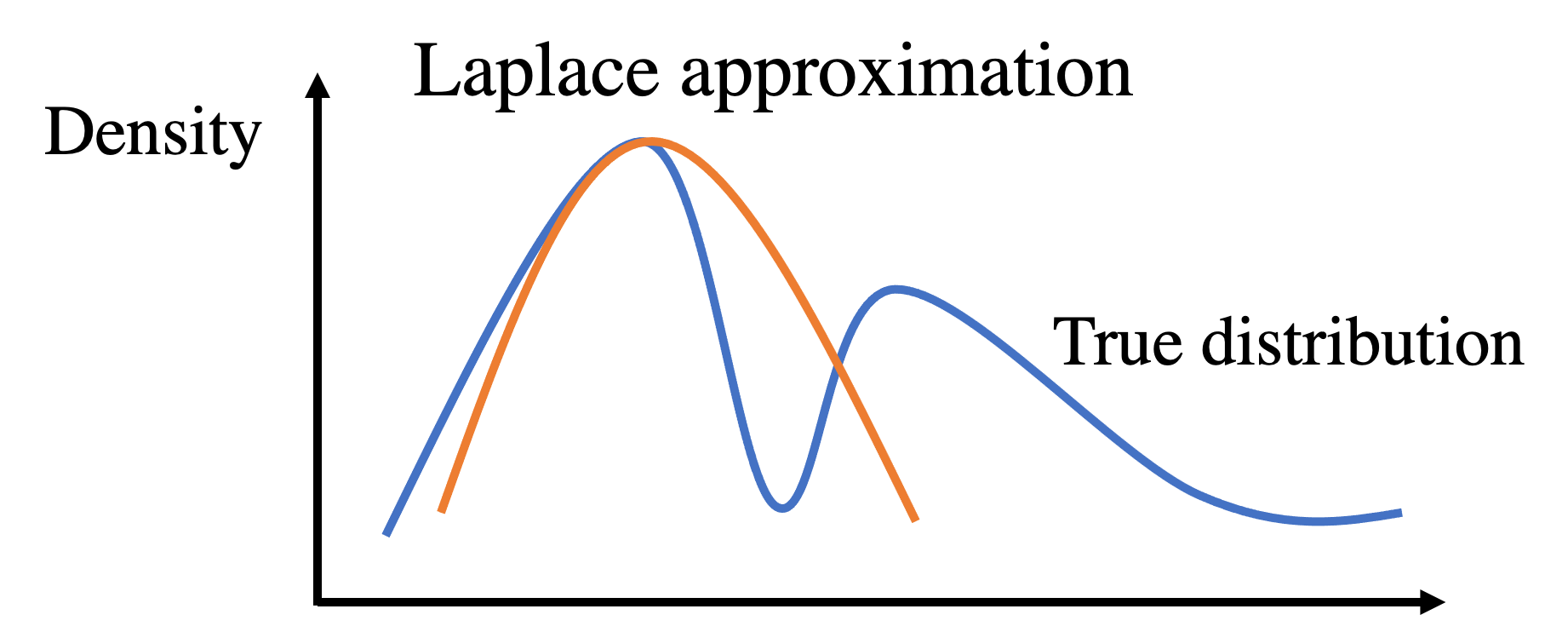}} \ 
    \subfloat[MC dropout]{\includegraphics[height = 1.3in]{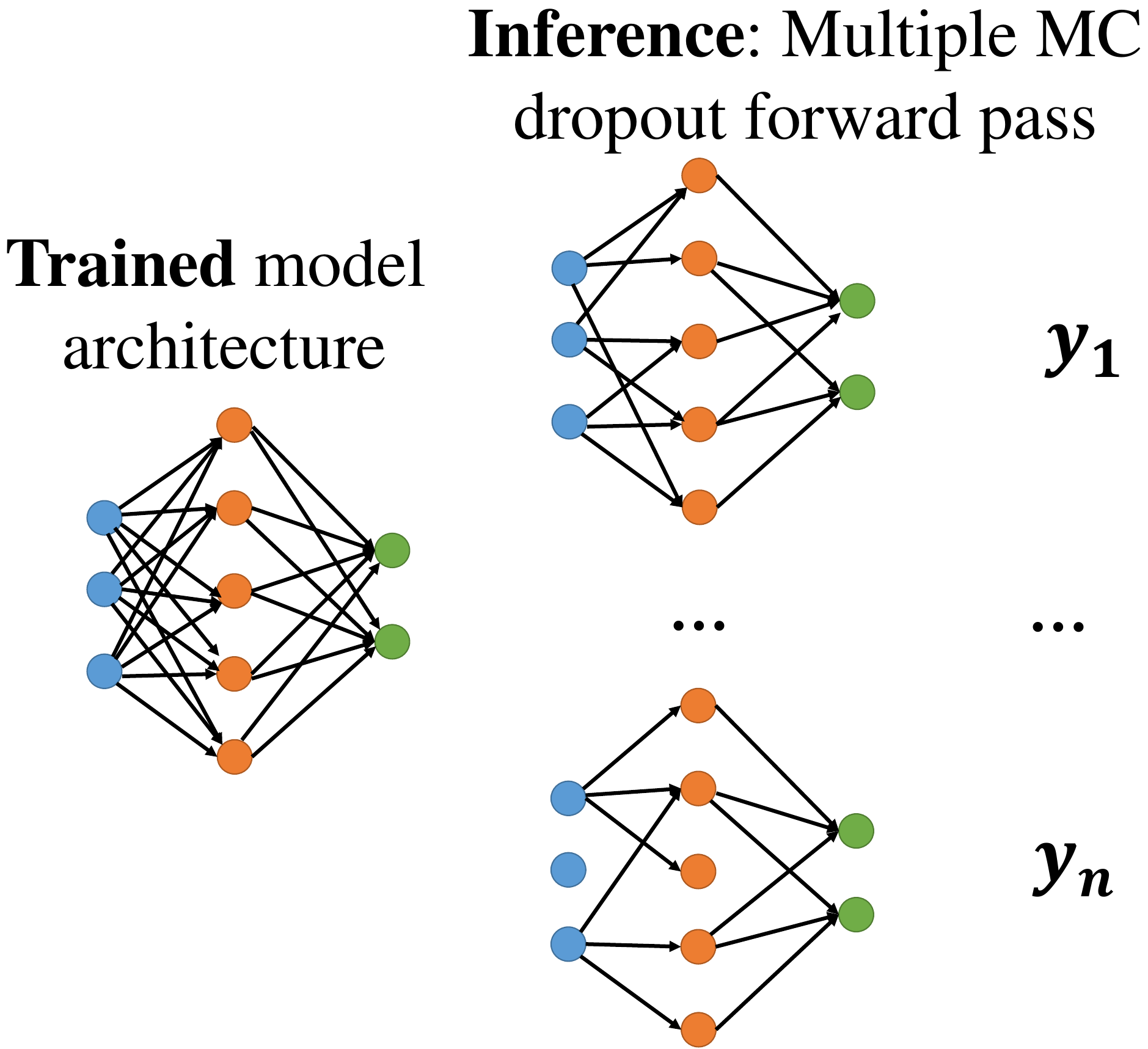}}
    \caption{Illustration of model uncertainty methods.}
    \label{fig:laplace}
\end{figure}

    

Moreover, for DNN, it is still infeasible to compute the inverted Hessian matrix for all parameters, which is typically in the order of several million.  Previous work on using Laplace approximation for neural network uncertainty quantification mainly aims to leverage the techniques of Hessian approximation to simplify the computation. One straightforward solution is to ignore the covariance between weights and only use the diagonal of the Hessian matrix. Inspired by the \textit{ Kronecker factored approximations} of the curvature of a neural network, the approach \cite{ritter2018scalable} factorizes the Hessian matrix into the Kronecker product (matrix operation that results in a block matrix) of two smaller matrices. The method brings down the inversion cost of the Hessian matrix and can be scaled to deep convolutional networks and applied to Bayesian online learning efficiently \cite{ritter2018online}. However, the method introduces an additional assumption that each layer of the neural network is independent (ignoring covariance between layers), which might lead to an overestimation of the variance (uncertainty) in certain directions. 

To make the Laplace approximation allow for uncertainty quantification for contemporary DNN, another challenge is to calibrate the predictive uncertainty.  One standard practice is to tune the prior precision of the Gaussian prior on model weights \cite{ritter2018scalable}. This has a regularizing effect both on the approximation to the true Hessian,  as well as the Laplace approximation itself, which may be placing probability mass in low probability areas of the true posterior. However, these parameters require optimization w.r.t. the prediction uncertainty performance on a validation dataset, which may not generalize well on new test data. To overcome this disadvantage, another approach introduces a more flexible framework to tune the uncertainty of Laplace-approximated BNNs by adding some additional '\textit{hidden units}' to the hidden layer of MLP-based network \cite{kristiadi2021learnable}. The framework is trained with an uncertainty-aware objective to improve the uncertainty calibration of Laplace approximations. 
However, the limitation of this method is it can only be applied to MLP-trained networks, and cannot generalize to other models, e.g., convolutional neural networks.

To sum up, the previous methods employed optimization-based schemes like variational inference and Laplace approximations of the posterior. In doing so, strong assumptions and restrictions on the form of the posterior are enforced. The common practice is to approximate the posterior with a Gaussian distribution. The restrictions placed are often credited with inaccuracies induced in predictions and uncertainty quantification performance. The difference between the two approximations is that the Laplace approximation is a local approximation around the MAP estimation, and it can be applied to a pre-trained neural network and obtain uncertainty quantification without influencing the performance of the neural network. On the other hand, the variational inference is a global optimization used during training and may influence the model prediction performance. 


 {\bf Markov Chain Monte Carlo approximation}:  Markov Chain Monte Carlo (MCMC) is a general method for sampling from an intractable distribution. MCMC constructs an ergodic Markov chain whose stationary distribution is posterior $p(\boldsymbol{\theta}|\mathbf{X}, \mathbf{Y})$. Then we can sample from the stationary distribution.
  The inference step of BNN   can be approximated as the following equation: 
  
\begin{equation}\label{eq:mcmc}\footnotesize
    p(y^*|\boldsymbol{x}^*,\mathbf{X}, \mathbf{Y} ) = \int p(y^*|\boldsymbol{x}^*, \boldsymbol{\theta})p(\boldsymbol{\theta}|\mathbf{X}, \mathbf{Y}) d\boldsymbol{\theta} \approx \frac{1}{N}\sum_{i=1}^{N}p(y^*|\boldsymbol{x}^*, \boldsymbol{\theta}_i).
\end{equation}

where $\boldsymbol{\theta}_i \sim p(\boldsymbol{\theta}|\mathbf{X}, \mathbf{Y})$ is a sample generated from MCMC. This process generates samples by a Markov chain over a state space, where each sample only depends on the state of the previous sample. The dependency is described with a proposal distribution  $T(\boldsymbol{\theta}'|\boldsymbol{\theta})$ that specifies the probability density of transitioning to a new sample $\boldsymbol{\theta}'$ from a given sample $\boldsymbol{\theta}$. Some acceptance criteria are based on the relative density (energy) of two successive samples evaluated at the posterior to determine whether to accept the new sample or the previous sample.  The vanilla implementation is through the \textit{Metropolis-Hastings} algorithm \cite{murphy2012machine} with Gaussian proposal distribution $T(\boldsymbol{\theta}'|\boldsymbol{\theta}) \sim \mathcal{N}(\boldsymbol{\theta}, \Sigma)$. Specifically, in each iteration, the algorithm constructs a Markov chain over the state space of $\boldsymbol{\theta}$ with the proposal density distribution. The proposal sample is stochastically accepted with the acceptance probability $\alpha(\boldsymbol{\theta}', \boldsymbol{\theta}) = \frac{T(\boldsymbol{\theta}'|\boldsymbol{\theta})p(\boldsymbol{\theta}'|\boldsymbol{x})}{T(\boldsymbol{\theta}|\boldsymbol{\theta}')p(\boldsymbol{\theta}|\boldsymbol{x})}$ via a random variable $\mu$ drawn uniformly from the interval $[0, 1]$ ($\mu \sim \text{Unif}(0, 1)$).  If we reject the proposal sample, we retain the previous sample $\boldsymbol{\theta}$.
This strategy ensures the stationary distribution of the samples converges to the true posterior after a sufficient number of iterations.
However,  the isotropic Gaussian proposal distribution shows random walk behavior and can cause slow exploration of the sample space and a high rejection rate. Thus it takes a longer time to converge to the stationary distribution.
The problem is more severe because of the high dimensional parameter space of modern DNN which hinders the application of MCMC 
on neural network parameter sampling \cite{neal2012bayesian}.

The recent development on MCMC for modern DNN mainly focuses on how to sample more efficiently and reduce convergence iterations. For example, one direction aims to combine the MCMC sampling with variational inference \cite{salimans2015markov,wolf2016variational} to take advantage of both methods. Since variational inference approximates the posterior by formulating an optimization problem w.r.t. the variational posterior is selected from a fixed family of distributions.  This approach could be fast by ensuring some constraints on the variational posterior format, but they may approximate the true posterior poorly even with very low ELBO loss. On the other hand, MCMC does not have any constraints on the approximated posterior shape, and can potentially approximate the exact posterior arbitrarily well with a sufficient number of iterations. 
\textit{Markov Chain Variational Inference} (MCVI) bridges the accuracy and speed gap between MCMC and VI by interpreting the iterative Markov chain $q(\boldsymbol{\theta}|\boldsymbol{x}) = q(\boldsymbol{\theta}_0)\prod_{t=1}^{T}q(\boldsymbol{\theta}_t|\boldsymbol{\theta}_{t-1}, \boldsymbol{x})$ as a variational approximation in the expanded space with $\boldsymbol{\theta}_0, ...\boldsymbol{\theta}_{T-1}$. Thus, instead of constructing a sequential Markov Chain to sample $\boldsymbol{\theta}_0,...\boldsymbol{\theta}_{T-1}$, they use another auxiliary variational inference distribution $r(\boldsymbol{\theta}_0,...,\boldsymbol{\theta}_{T-1})$ to approximate the true distribution with some flexible parametric form.  By optimizing the lower bound over the parameters of the variational distribution with a neural network, we can obtain the samples. 


The advantage of MCMC methods is that the samples it gives are guaranteed to converge to the exact posterior after a sufficient number of iterations. This property allows us to control the trade-off between sampling accuracy and computation. However, the downside of this method is that we do not know how many iterations are enough for convergence and it may take an excessive amount of time and computing resources.

{\bf Monte-Carlo (MC) dropout}: The MC dropout approach  \cite{gal2016dropout} is currently among the most popular methods for DNN uncertainty quantification due to its simplicity and ease of implementation. The main idea is that the optimization of a neural network with a dropout layer can be equivalent to approximating a BNN with variational inference on a parametric Bernoulli distribution \cite{gal2016dropout}. Uncertainty estimation can be obtained by computing the variance of multiple stochastic forward predictions with different dropout masks (switching off some neurons' activations). The average predictions with various weights dropout can be interpreted as approximating the integration over the model's weights (as Eq.\ref{eq:bayesian}) whose variational distribution follows the Bernoulli distribution.   MC dropout offers several advantages. First, it requires minimal modification to the existing DNN architecture design, allowing for straightforward implementation in practice. Second, it addresses the problem of representing uncertainty without sacrificing model accuracy. Uncertainty is added only at inference time through stochastic sampling rather than by modifying the learning objective. However, although there is theoretical intuition for the probabilistic interpretation of MC dropout from a variational approximation perspective, MC dropout tends to be less calibrated than other UQ methods on many uncertainty benchmark datasets \cite{guo2017calibration}.

In summary, we review several approximation methods for BNNs designed to reduce computational and memory demands. These methods capture model uncertainty associated with the parameters. However, the need for approximation may lead to less accurate uncertainty estimates, and in practice, these methods remain computationally intensive.

\subsubsection{Ensemble  models}
Ensemble models combine multiple neural networks to form an output distribution, where the variability of the distribution quantifies model uncertainty. {\color{black} To capture model uncertainty from various sources, several strategies for constructing ensembles have been adopted. The first strategy involves bootstrapping \cite{lakshminarayanan2017simple,laurentpacked}. This approach involves random sampling from the original dataset with replacement. An ensemble of neural networks is then constructed, with each model trained on different bootstrapped samples. After training, inference is performed by aggregating the ensemble predictions, with uncertainty obtained from the prediction variance (for regression) or average entropy (for classification). The second strategy is to construct different neural network architectures by varying the number of layers, hidden neurons, and types of activation functions \cite{mallick2022deep,wild2024rigorous}. This strategy can account for the uncertainty from model misspecification. Other strategies involve different parameter initializations and random shuffling of datasets. This approach is better than the bootstrap strategy since more samples can be utilized for each model.  The third type is the hyperensemble approach  \cite{wenzel2020hyperparameter}. This approach constructs ensembles with different hyper-parameters, such as learning rate, optimization strategy, and training strategy.

Although ensemble models are relatively simple to implement, they have several limitations. First, they have a high computational cost as they require training multiple independent networks and maintaining all networks in memory during inference. Second,  {model diversity} is required to ensure accurate uncertainty estimation.

}

\subsubsection{ Sample distribution-related methods}
In Section 3.1.1, we described model uncertainty related to sample distribution as a distinct category. It further includes two cases: (1) test samples and training samples follow different distributions (out of distribution); (2) a test sample is far from other training samples (or is surrounded by sparse training samples) in the feature space. Here, we focus only on methods for case (2), i.e., sample density (distance)-aware neural networks, and do not specifically discuss out-of-distribution (OOD) methods, as BNNs and ensemble methods can also be used for OOD-related uncertainty. Methods for addressing OOD problems will be reviewed in Section \ref{sec:mlproblems}.
Existing methods can be grouped into two categories: Gaussian process hybrid neural network and distance-aware neural network. 

 {\color{black} \bf Background of Gaussian Process:}
A Gaussian process (GP) is a type of stochastic process where any finite collection of random variables follows a multivariate Gaussian distribution \cite{williams2006gaussian}.
    Given a set of points $\{\boldsymbol{x}_i\}_{i=1}^n$, a GP defines a prior over functions $y_i = f(\boldsymbol{x}_i)$, and assumes the $p(y_1, ..., y_n)$ follows the Gaussian distribution $\mathcal{N}(\boldsymbol{\mu}(\boldsymbol{x}), \boldsymbol{\Sigma}(\boldsymbol{x}))$, where $\boldsymbol{x}=(\boldsymbol{x}_1,...,\boldsymbol{x}_n)$, $\boldsymbol{\mu}(\boldsymbol{x})$ is the mean function, and $\boldsymbol{\Sigma}(\boldsymbol{x})$ is the covariance function based on $\boldsymbol{\Sigma}_{ij} = \kappa(\boldsymbol{x}_i, \boldsymbol{x}_j)$.  $\kappa$ is a positive definite kernel function (e.g., radial basis function) that measures the similarity between pairs of input samples and controls the smoothness of the GP.   For GP inference, given a new sample $\boldsymbol{x^{*}}$, the joint distribution between prediction for the new sample $y^{*}$ and the target variable $\mathbf{y}$ of the training samples is shown in Eq.~\ref{eq:GPeq}, where $\mathbf{K}_{n}$ {is the covariance matrix between the $n$ training samples}, $\mathbf{K}_{\boldsymbol{x}}$ is the covariance between the test sample and training samples, and $K^{*}$ is the prior variance of the test sample.   
    \begin{equation}\label{eq:GPeq}
    \footnotesize
        \begin{pmatrix}
        \mathbf{y} \\
        y^{*} 
        \end{pmatrix} = \mathcal{N}\Bigg( \begin{pmatrix}
        \boldsymbol{\mu} \\
        \mu^{*} 
        \end{pmatrix}, \begin{pmatrix}
        \mathbf{K}_{n} & \mathbf{K}_{\boldsymbol{x}} \\
        \mathbf{K}_{\boldsymbol{x}}^{T} & K^{*} 
        \end{pmatrix} \Bigg).
    \end{equation}
    \begin{figure}
        \centering
        \includegraphics[height=1.3in]{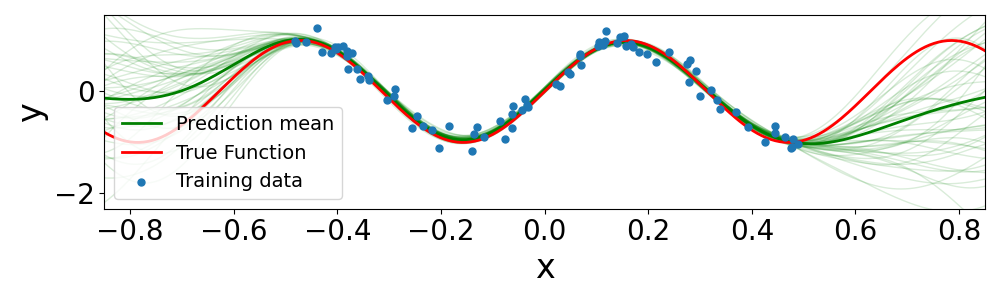}
        \caption{Gaussian Process inference example: green lines are the prediction sample distribution.}
        \label{fig:gpvis}
    \end{figure}
The prediction for a new test sample $\boldsymbol{x}^{*}$ is obtained by computing the posterior distribution conditioned on the training data $\mathcal{D}_{\text{tr}}$, given by Eq.~\ref{eq:gpposterior}. 
   GP inference yields lower uncertainty when test samples are in regions where training samples are abundant (higher sample density as shown in the middle part of Fig.~\ref{fig:gpvis}), otherwise resulting in higher uncertainty (boundary part of Fig.~\ref{fig:gpvis}).  Although GP is fundamentally a Bayesian method, it differs from BNNs by primarily capturing uncertainty related to sample sparsity, rather than uncertainty over model parameters.  Therefore, we categorize GP as a separate sub-category of UQ methods. 
{\color{black}   \begin{equation}\label{eq:gpposterior}\footnotesize
       p(y^{*}|\boldsymbol{x}^{*}, \mathcal{D}_{\text{train}}, \boldsymbol{\theta}) = \mathcal{N}(y|  \mathbf{K}_{\boldsymbol{x}}^T\mathbf{K}_{n}^{-1}\mathbf{y},   K_{\boldsymbol{x}^*} - \mathbf{K}_{\boldsymbol{x}}^T\mathbf{K}_{n}^{-1}\mathbf{K}_{\boldsymbol{x}}).
   \end{equation}
   }
 {\textit{Sparse Gaussian process:} } 
Though GP has a sound theoretical framework for uncertainty estimation,  it is computationally unscalable for large datasets because inverting the covariance matrix requires $\mathcal{O}(n^3)$ time complexity ($n$ is the total number of training samples). To mitigate this bottleneck, many methods \cite{snelson2005sparse,titsias2009variational} attempt to make a sparse approximation to the full GP to reduce the computational complexity to $\mathcal{O}(m^2n)$ ($m$ is the number of inducing variables, and $m \ll n $). The inducing variables can be anywhere in the input domain, and are not constrained to be a subset of the training data and are represented asinput-output pairs $\{\hat{\boldsymbol{x}}_i, \hat{y}_i\}_{i=1}^m$. Thus, inverting the original covariance matrix $\mathbf{K}_n$ can be replaced with a low-rank approximation from the inducing variables, which only requires the inversion of an $m\times m$ matrix $\mathbf{K}_m$. Then, the question becomes how to select the $m$ best-inducing variables to be representative of the training dataset. Common approaches assume that the best representative inducing variables are those that maximize the likelihood of the training dataset \cite{snelson2005sparse}. Subsequently, the location of inducing variables and the hyper-parameters of GP are optimized simultaneously through maximum likelihood. The training data likelihood can be obtained by marginalizing the inducing variables on the joint distribution of the  training dataset and inducing variables. 


 Besides the computational challenge, another limitation of GP is that the joint  Gaussian distribution assumption on the target variables limits the model's capability to capture diverse relationships among instances within large datasets. Additionally, GP relies heavily on the kernel function to compute the similarity between samples by transforming input features into a high-dimensional manifold. However, for high-dimensional structured data, it is challenging to construct appropriate kernel functions to extract hierarchical features for computing similarity between samples. To address these limitations, two research areas have been proposed: \textit{deep kernel learning} and \textit{Deep (Compositional) Gaussian Process}. 

{\color{black} \bf{Gaussian Process Hybrid Neural Network}}
 
 \textit{Deep kernel learning} \cite{wilson2016deep} aims to combine the structured feature learning capability of DNN with a GP to learn more flexible representations. The motivation is that DNN can automatically discover meaningful representations from high-dimensional data, which could alleviate the fixed kernel limitations of GP and improve its expressiveness. Specifically, the deep kernel learning approach transforms the kernel $\mathbf{K}_{\boldsymbol{\theta}}(\boldsymbol{x}_i, \boldsymbol{x}_j)$ to  $\mathbf{K}_{\boldsymbol{\theta}}(g(\boldsymbol{x}_i; \boldsymbol{w}), g(\boldsymbol{x}_j, \boldsymbol{w})) $, where $g(\cdot;\boldsymbol{w})$ is the neural network parameterized with $\boldsymbol{w}$ and $\mathbf{K}_{\boldsymbol{\theta}}$ is the base kernel function (e.g., radial basis function) of GP.  Deep learning transformation can capture the non-linear and hierarchical structure in high-dimensional data. The  GP with the base kernel is applied on the final layer of DNN and makes inferences based on the learned latent features, as shown in Fig.~\ref{fig:gp} (a). The idea has been successfully applied to spatio-temporal crop yield prediction, where GP plays a role in accounting for the spatio-temporalautocorrelation between samples \cite{you2017deep}, which may not be captured by the DNN features alone. 
 

\begin{figure}[ht]
    \centering
    \subfloat[Deep kernel learning]{\includegraphics[height = 1.0in]{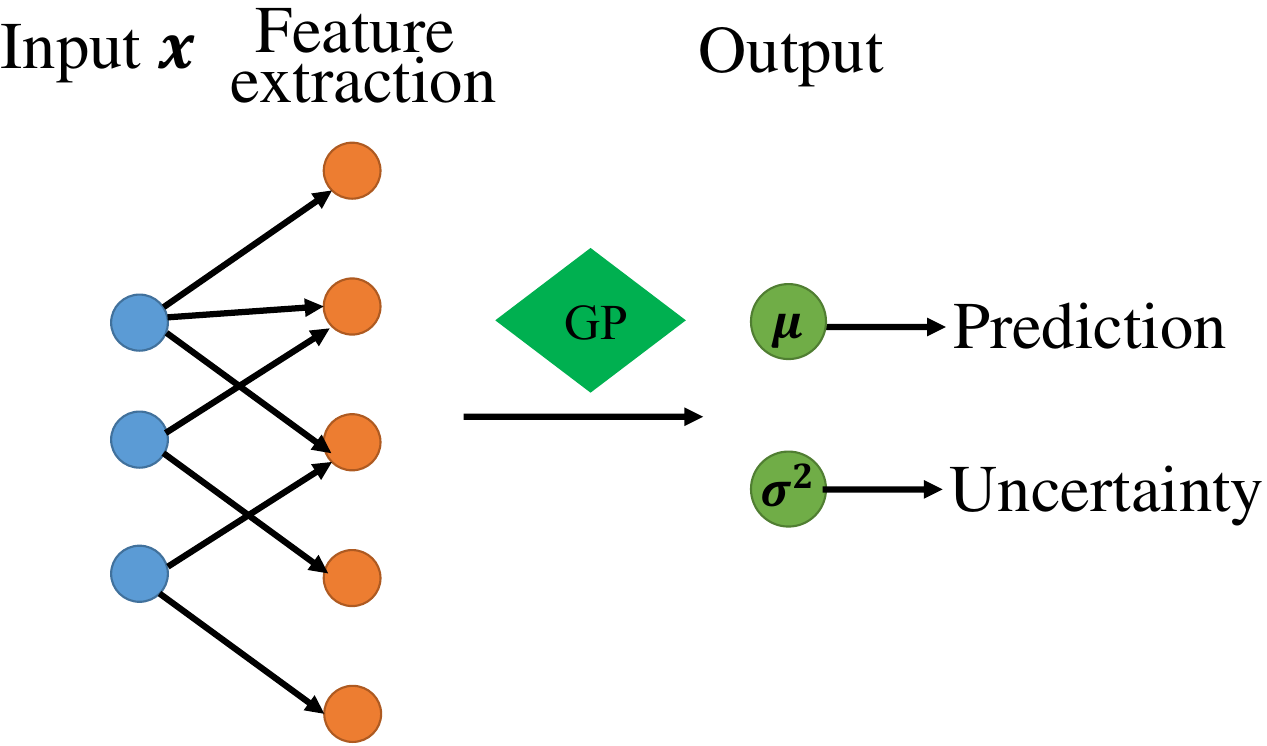}} 
    \hspace{0.8in}
    \subfloat[Deep Gaussian process]{\includegraphics[height = 0.7in]{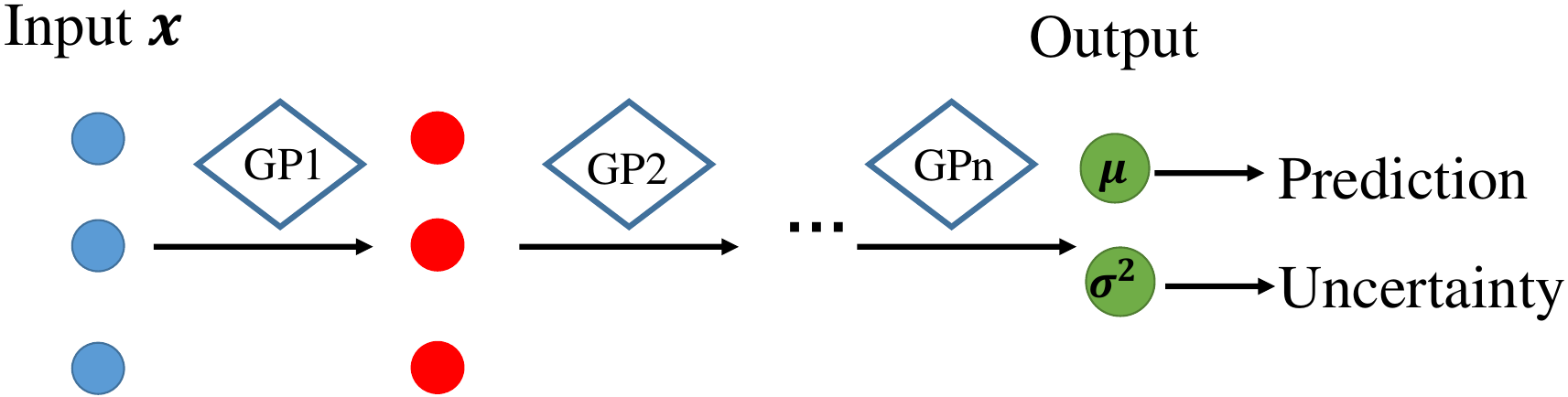}}
    \caption{\color{black} Illustration of deep kernel learning and deep (compositional) Gaussian process.}
    \label{fig:gp}
\end{figure}

 \textit{Deep (compositional) Gaussian process} \cite{damianou2015deep} represents another category, focusing on function composition inspired by deep neural network architecture. In this model, each layer is a GP model whose inputs are determined by the output of the preceding GP, as shown in Fig.~\ref{fig:gp} (b). The recursive composition of GPs results in a more complex distribution over the predicted target variables, which addresses the joint Gaussian distribution limitation of traditional GP. The forward propagation and joint probability distribution of the model can be expressed as follows:\begin{equation}\footnotesize
\begin{split}
    & \boldsymbol{y} = \boldsymbol{f}_L(\boldsymbol{f}_{L-1}(...\boldsymbol{f}_1(\boldsymbol{x})))  
    \ \ \text{and} \ \  p(y, \boldsymbol{f}_L,...\boldsymbol{f}_1|\boldsymbol{x}) \sim p(\boldsymbol{y|\boldsymbol{f}_L})\prod_{i=2}^{L}p(\boldsymbol{f}_i|\boldsymbol{f}_{i-1})p(\boldsymbol{f}_1|\boldsymbol{x})
\end{split},
\end{equation}
where each function $\boldsymbol{f}_i(\cdot)$ represents a Gaussian process model. The intermediate distributions follow  Gaussian distributions, but the final distribution will capture a more complex distribution over the target variable $\boldsymbol{y}$. The composition also allows uncertainty to propagate from the input through each intermediate layer. However, the challenge associated with the compositional Gaussian process lies in maximizing the data likelihood $p(\boldsymbol{y}|\boldsymbol{x}) $; the direct marginalization of hidden variables $\boldsymbol{f}_i$ is intractable. To overcome this challenge, variational inference introduces inducing points on each hidden layer and by optimizing over the variational distribution $q(\boldsymbol{f}_i)$. Then, the marginal likelihood lower bound can be obtained by propagating the variational approximation at each layer \cite{ustyuzhaninov2020compositional}. The framework also allows for incorporating partial or uncertain observations into the model by placing a prior over the input variables $\boldsymbol{x}$ and propagating uncertainty layer by layer \cite{damianou2016variational}. 



\begin{table*}[ht]
\footnotesize
\centering
\caption{\color{black} Comparison of UQ methods for model uncertainty. }
\label{tab:modeluq}
\begin{tabular}{|l|l|l|l|}
\hline
\textbf{Category} & \textbf{Method} & \textbf{Pros} & \textbf{Cons} \\ \hline
\color{black}
\multirow{3}{*}{\begin{tabular}[c]{@{}l@{}}\textbf{BNN}: Capture parameter \\uncertainty via posterior \\ estimation.\end{tabular}} 
& \begin{tabular}[c]{@{}l@{}}Variational Inference \&\\ Laplace Approximation\\ \cite{posch2019variational,louizos2017multiplicative,ritter2018scalable,ritter2018online,kristiadi2021learnable} \end{tabular} 
& \begin{tabular}[c]{@{}l@{}} Practically efficient \\ for  large models\end{tabular} 
& \begin{tabular}[c]{@{}l@{}}Approximation is \\ based on assumptions \\ (e.g., Gaussian distribution)\end{tabular} \\ \cline{2-4}

& \begin{tabular}[c]{@{}l@{}}MCMC\\ \cite{neal2012bayesian,salimans2015markov,wolf2016variational}\end{tabular} 
& \begin{tabular}[c]{@{}l@{}} Flexible for any \\ distribution assumption \end{tabular}
& High computational cost \\ \cline{2-4}

& \begin{tabular}[c]{@{}l@{}}MC Dropout\\ \cite{gal2016dropout,guo2017calibration}\end{tabular} 
& \begin{tabular}[c]{@{}l@{}} Simple, scalable, flexible \\ for large neural networks \end{tabular}
& Lacks theoretical grounding \\ \hline

\begin{tabular}[c]{@{}l@{}}\textbf{Ensemble}: Capture  \\ uncertainty from models, \\ parameters, and \\ hyperparameters\end{tabular} 
& \begin{tabular}[c]{@{}l@{}}Network/Bootstrap \\ /Hyper-Ensemble\\ \cite{mallick2022deep,lakshminarayanan2017simple,wenzel2020hyperparameter} \end{tabular} 
& \begin{tabular}[c]{@{}l@{}}Capture uncertainty \\ from architecture, learning \\ algorithms, hyperparameters\end{tabular} 
& High computational cost \\ \hline

\multirow{2}{*}{\begin{tabular}[c]{@{}l@{}}\textbf{Sample Distribution}:\\ Uncertainty due to \\ OOD inputs.\end{tabular}} 
& \begin{tabular}[c]{@{}l@{}}Deep Gaussian Processes\\ \cite{wilson2016deep,you2017deep,damianou2015deep}\end{tabular} 
&\begin{tabular}[c]{@{}l@{}}Strong theoretical \\ grounding in GPs \end{tabular} 
&\begin{tabular}[c]{@{}l@{}}Poor scalability \\  to large datasets \end{tabular}  \\ \cline{2-4}

& \begin{tabular}[c]{@{}l@{}}Distance-aware DNNs\\ \cite{liu2020simple,van2020uncertainty,van2021on}\end{tabular} 
& Simple and efficient 
& \begin{tabular}[c]{@{}l@{}}Embedding distances may \\  not reflect sample similarity\end{tabular}  \\ \hline
\end{tabular}
\end{table*}

 {\bf Distance-aware neural network:} Although modern neural networks can  extract representative 
 \begin{wrapfigure}{r}{0.60\textwidth}   
  \vspace{-4mm}
  \centering
      \includegraphics[width=1.6in]{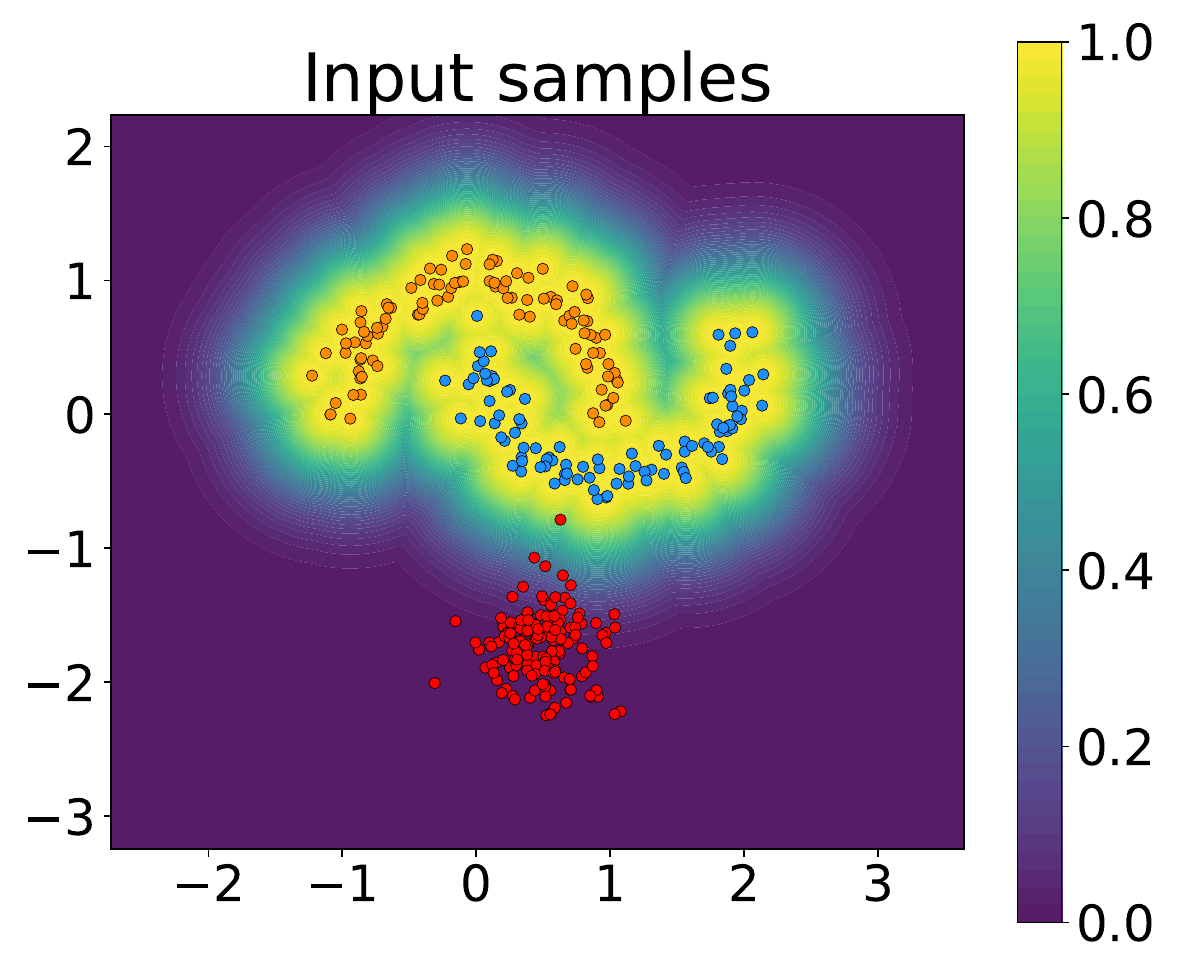}
      \includegraphics[width=1.5in]{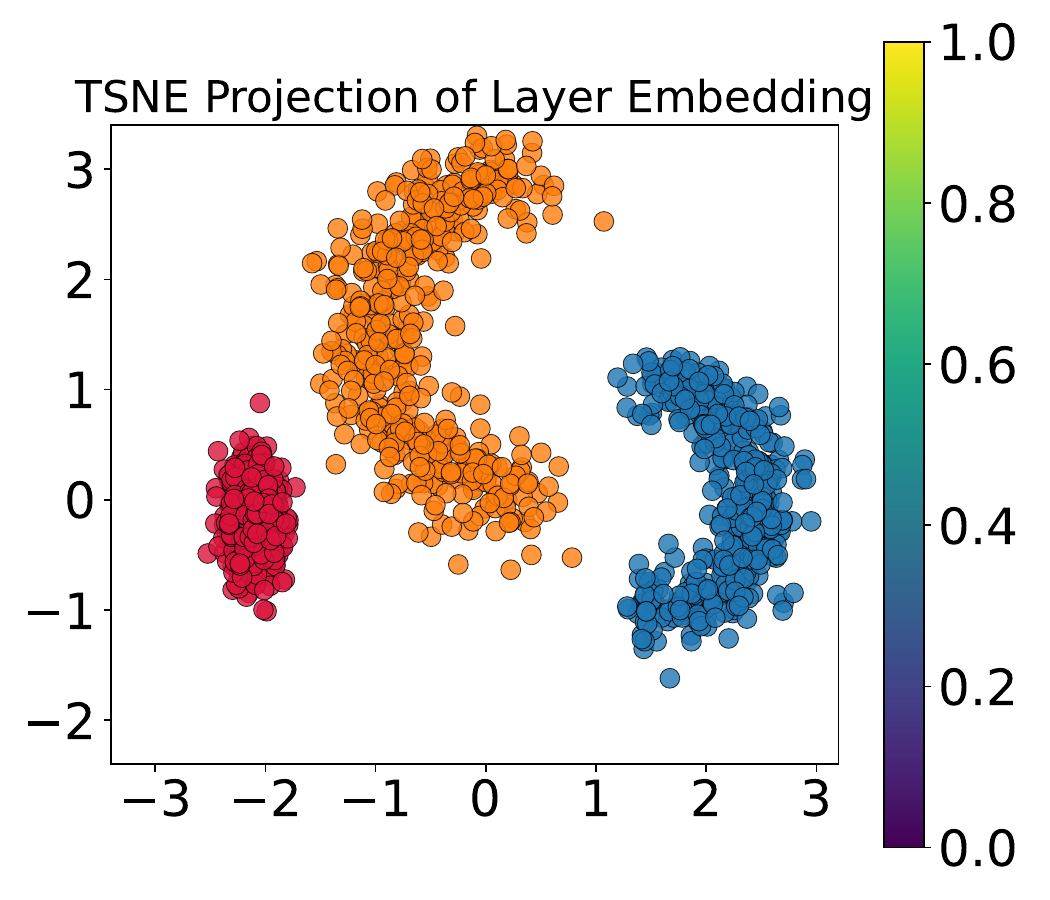}
  \vspace{-2mm}
  \caption{Two variants of the architecture (adapted from
           \cite{liu2020simple}).}
  \label{fig:distance-awarenn}
\end{wrapfigure}
features from large datasets, they do not consider how distinct new test samples may be from the training dataset. To address the uncertainty resulting from sample feature density, many approaches incorporate distance awareness between samples into neural network design, inspired by Gaussian processes \cite{liu2020simple}. Assume the input data manifold is equipped with a metric $||\cdot||_{\mathcal{X}}$, quantifying the distance between samples in the feature space.   The intuition behind a distance-aware neural network is to leverage DNNs' feature extraction capability to learn a hidden representation $h(\boldsymbol{x})$ that reflects a meaningful distance in the data manifold $||\boldsymbol{x} - \boldsymbol{x'}||_{\mathcal{X}}$ as shown in Fig~\ref{fig:distance-awarenn}.
    However, one significant issue with the unconstrained DNN model is  the \textit{feature collapse}, which means DNN feature extraction can map \textit{in-distribution} data (training samples) and \textit{out-of-distribution} data (which lies further from the training data)
    to similar latent representations. Thus, the Gaussian process based on the DNN extracted feature can be over-confident for those samples that lie further away from training samples.     
    To avoid the feature collapse problem, several constraints have been proposed: sensitivity and smoothness \cite{van2020uncertainty,van2021on}. Sensitivity implies that a small change in the input should result in a small change in the feature representation,  which helps ensure distinct samples are mapped to different latent features. Smoothness implies that small changes in the input should not cause dramatic changes in the output. In general, these two constraints 
    can be ensured by  \textit{bi-Lipschitz} constraints \cite{liu2022simple}, which means the relative changes in the hidden feature representation $h_{\boldsymbol{\theta}}(\boldsymbol{x})$ are bounded by changes in input space as Eq.~\ref{eq:bilip} shows, where $L_1$ and $L_2$ are constants.
    \begin{equation}\label{eq:bilip}\footnotesize
L_1*|\boldsymbol{x} - \boldsymbol{x}'|_{\mathcal{X}}    <    |h_{\boldsymbol{\theta}}(\boldsymbol{x}) - h_{\boldsymbol{\theta}}(\boldsymbol{x}')|_{\mathcal{H}} < L_2*|\boldsymbol{x} - \boldsymbol{x}'|_{\mathcal{X}}.
    \end{equation}
    To enforce bi-Lipschitz constraints to DNN, two approaches have been proposed: spectral normalization and gradient penalty. Spectral normalization \cite{liu2020simple} claims that the bi-Lipschitz constants $L$ can be ensured to be less than one by normalizing the weights matrix in each layer with the spectral norm. This method is fast and effective for practical implementation. The other approach is called gradient penalty \cite{van2020uncertainty}, which introduces another loss penalty: the square gradient at each input sample $\nabla_{\boldsymbol{x}}^2h_{\boldsymbol{\theta}}(\boldsymbol{x})$. This will add a soft constraint to the neural networks to constrain the Lipschitz coefficients. Gradient penalty is a soft constraint compared to spectral normalization and is computationally more intensive. Some works extend the distance-aware framework to the non-parametric estimation of the conditional label distribution, enabling more flexible modeling of the distribution \cite{kotelevskii2022nonparametric,wang2023gaussian}. 
    
{\color{black}
{\bf Summary of Model Uncertainty:}
 We summarize and compare existing methods for quantifying model uncertainty in Table~\ref{tab:modeluq}. BNN models can capture model uncertainty arising from parameter estimation but usually have very high computational costs, making them infeasible for practical applications. The ensemble models can capture uncertainty from multiple perspectives, such as model architecture misspecification, limited training dataset, and choices of hyperparameters. The method also has a high computational cost. On the other hand, the sample distribution-based model can capture uncertainty due to distribution shifts, but it's often hard to learn the distance-aware feature space and the method requires adding constraints to the neural network model.
 }

{\color{black}Although we categorize BNN and ensemble methods under model uncertainty, they can be used to capture both model and data uncertainty simultaneously with minor modification. For BNNs, as shown in Equation~\ref{eq:bayesian}, model uncertainty is reflected by the posterior distribution over model parameters, while data uncertainty can be captured by averaging the predictive distributions from multiple model samples. Similarly, for ensemble methods, data uncertainty can be further obtained through averaging the class probability predictions across individual models. A common approach to obtain the class probability is through the softmax of output logits. However, this method may lead to overconfident predictions~\cite{guo2017calibration}. To address this issue,  a variety of calibration techniques, including both parametric and non-parametric approaches, have been developed to capture the predicted distribution for data uncertainty. More detailed discussions are in Section~\ref{subsec:combinedatamodel}.}

\subsection{Data Uncertainty}
This section discusses the existing methodologies that quantify  data uncertainty in DNN models. Data uncertainty is modeled by the distribution $p(y|\boldsymbol{x}, \boldsymbol{\theta})$, where $\boldsymbol{\theta}$ represents the neural network parameters. We categorize these approaches into deep discriminative models and deep generative models.

\subsubsection{Deep discriminative model }
To quantify data uncertainty, a discriminative model directly outputs a predictive distribution using a neural network. Specifically, the distribution can be learned by a parametric or non-parametric model. A parametric model assumes that the output follows a specified family of probability distributions with parameters (e.g., mean and variance for a Gaussian distribution) estimated by a neural network. In contrast, the non-parametric model does not have any assumption on the underlying distributions. We will discuss existing methods for each category in detail.

    \textbf{Parametric model}: The standard approach for quantifying data uncertainty is to directly learn a parametric model for $p(y|\boldsymbol{x}, \boldsymbol{\theta})$. From a frequentist perspective, there exists a single set of optimal parameters $\boldsymbol{\theta}^*$.  For the classification problem, $p(y|\boldsymbol{x}, \boldsymbol{\theta})$ is a parameterized categorical distribution over $k$ classes, 
    with distribution parameters $\boldsymbol{\pi}=(\pi_1,...,\pi_k)$ predicted by the model  output as shown in Eq.~\ref{eq:categorical}. 
    \begin{equation}\label{eq:categorical}\footnotesize
        p(y|\boldsymbol{x}, \boldsymbol{\theta}) = \text{Categorical}(y;\boldsymbol{\boldsymbol{\pi}}) \ , \ \boldsymbol{\pi} = f(\boldsymbol{x}; \boldsymbol{\theta}), \ \sum_{c=1}^{k}\pi_c = 1, \ \pi_c > 0.
    \end{equation}
     In order to obtain the categorical distribution parameters,  a straightforward approach uses the softmax probability output $\pi_i = \frac{\exp(h_i(\boldsymbol{x}; \boldsymbol{\theta}))}{\sum_{c=1}^k\exp(h_c(\boldsymbol{x}; \boldsymbol{\theta}))}$ as the predicted uncertainty, but these methods tend to be over-confident because the softmax operation squeezes the prediction probability toward extreme values (zero or one) for the vast majority range of $h_i$  \cite{hendrycks2016baseline}. Subsequent work \cite{guo2017calibration} calibrate the softmax uncertainty with \textit{temperature scaling}, which simply adds an additional hyperparameter $T$ to the softmax calculation as $p = \frac{\exp(h_i(\boldsymbol{x})/T)}{\sum_{c=1}^k\exp(h_c(\boldsymbol{x})/T)}$ to overcome the overconfident outputs. This approach is straightforward to implement, but the result may still be overconfident due to a lack of constraints and requires calibration of the parameter.
    
     For regression problems, data uncertainty is assumed to arise from inherent noise in the training data (e.g., measurement or labeling error). In general, the training data is modeled as independent additive Gaussian noise with sample-dependent variance $\sigma(\boldsymbol{x})$, which indicates the target variable $y_i = f_{\boldsymbol{\theta}}(\boldsymbol{x}_i) + \epsilon(\boldsymbol{x}_i)$. $\epsilon(\boldsymbol{x}_i)$ is the independent heterogeneous Gaussian noise, representing each sample's uncertainty.
    In this way, the output will be a parameterized continuous Gaussian distribution, as Eq.~\ref{eq:gaussian} and Fig.~\ref{fig:pi} (a) show. The mean and variance are predicted from the neural network \cite{kendall2017uncertainties}, where the mean represents the model’s prediction, and the variance represents the uncertainty of each sample's prediction. To optimize the neural network parameters $\boldsymbol{\theta}$, maximum likelihood optimization is performed jointly on the mean and variance as Eq.~\ref{eq:gaussian} shows. This is also known as heteroscedastic regression, which assumes the observational noise level varies with different samples. This is suitable for cases where some samples have higher noise (uncertainty), while others have lower. Besides Gaussian distribution, the neural network can also be parameterized with other types of distributions, such as mixture Gaussian distribution \cite{guillaumes2017mixture}, which is implemented with mixture density network (MDN) \cite{bishop1994mixture}, assuming multiple modes for the prediction. MDN has the advantage of accounting for the uncertainty from multiple prediction modes but consumes more computation.  Choosing a suitable parameterized distribution is essential and depends on the problem.  
     {\color{black} \begin{equation}\label{eq:gaussian}\footnotesize
    \begin{split}
        &p(y|\boldsymbol{x}, \boldsymbol{\theta}) = \mathcal{N}(f_{\boldsymbol{\theta}}(\boldsymbol{x}), \sigma_{\boldsymbol{\theta}}(\boldsymbol{x})) \ \ \text{and} \ \
        \mathcal{L}_{\text{NN}}(\boldsymbol{\theta}) = \frac{1}{n}\sum_{i=1}^n\frac{1}{2\sigma_{\boldsymbol{\theta}}^2(\boldsymbol{x}_i)}|| y_i - f_{\boldsymbol{\theta}}(\boldsymbol{x_i})||^2 + \frac{1}{2}\log \sigma_{\boldsymbol{\theta}}^2(\boldsymbol{x}_i).
    \end{split}
    \end{equation}
    }
    The advantage of using a predictive distribution is that it can be easily incorporated into existing neural network architectures and requires slight modification to the training and inference process. However, the explicit parameterization form requires the model to choose an appropriate distribution to effectively capture the underlying uncertainty, which can be hard if no prior information is available.

    \textbf{Non-parametric model:} A widely popular approach for indicating data uncertainty is through \textit{prediction interval} (PI)  \cite{pearce2018high}. For regression problems, the prediction intervals output a lower and upper bound $[y_l, y_u]$, where we expect the ground truth $y$ falls within the interval with a prescribed confidence level,  $1-\alpha$, meaning that $p(y \in [y_l, y_u])> 1-\alpha$ as shown in Fig.~\ref{fig:pi} (b). 
    This approach is more flexible and does not require explicit distribution over the prediction variable.  Traditional prediction intervals are typically constructed in two steps: first, to learn the point estimation of the target variable, obtained by minimizing an error-based loss function (e.g., mean square loss), followed by estimating the prediction variance around the local optimum prediction. The strategy tries to minimize the prediction error but does not optimize the interval quality. {\color{black} One  approach~\cite{khosravi2010lower}} explicitly constructs a lower and upper bound estimation (LUBE) to directly optimize the PI characteristics, namely the width and coverage probability. The basic intuition is that the PI should cover the ground truth with a certain pre-defined probability (confidence level), but should be as narrow as possible. This approach enhances the quality of the constructed PI, but the resulting cost function is non-differentiable and requires Simulated Annealing (SA) sampling to obtain the optimal NN parameters. 
    

\begin{figure}
    \centering
    \subfloat[Prediction distribution example.]{\includegraphics[height=0.9in]{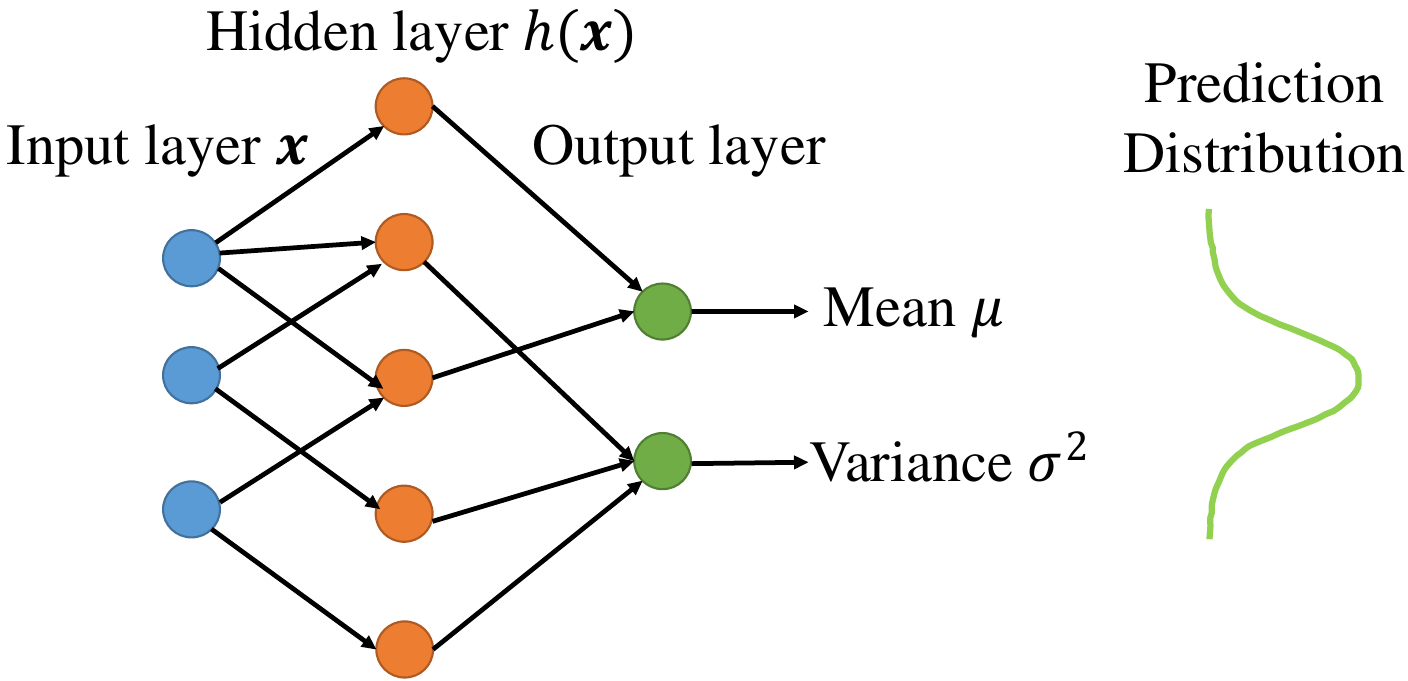}}  \hspace{7mm}
   \subfloat[Prediction interval example.]{\includegraphics[height=0.9in]{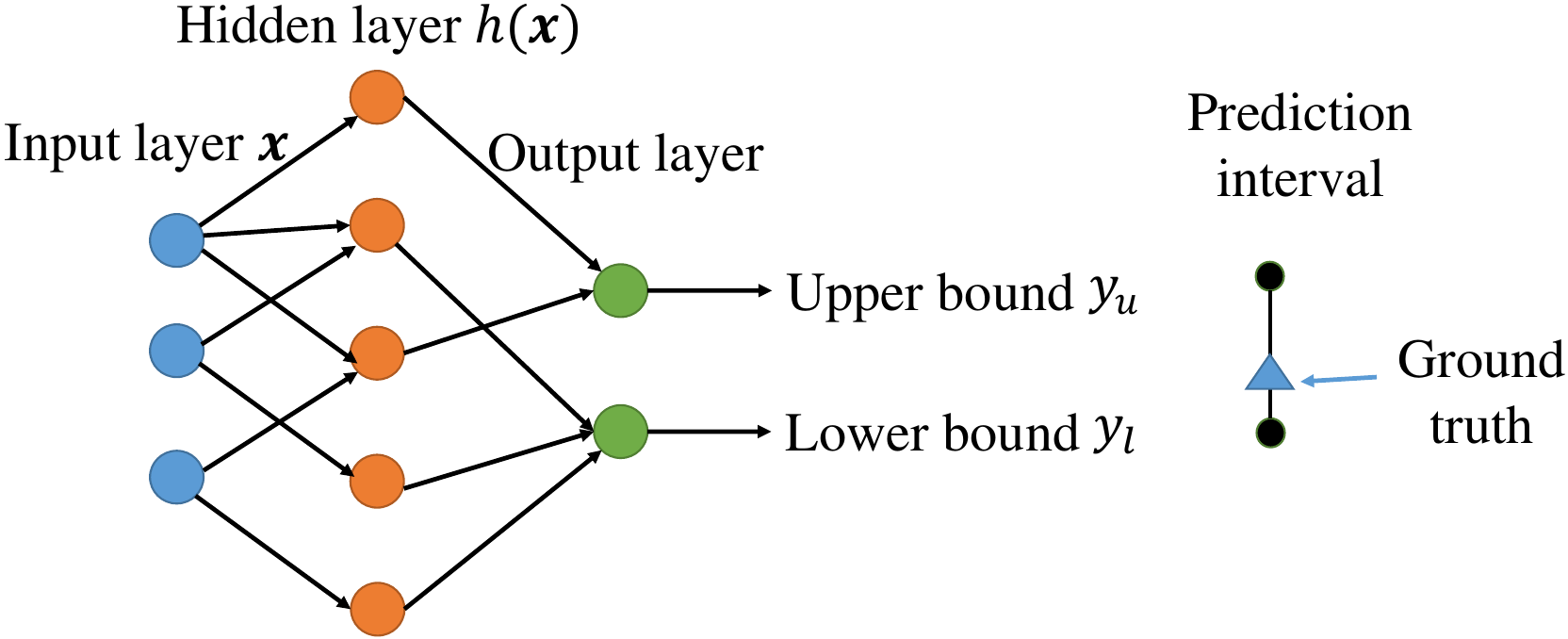}}
    \caption{Neural network architecture for the parametric and non-parametric model.}
    \label{fig:pi}
\end{figure}
    
    To address the non-differentiable limitation of LUBE, an alternative approach uses a coverage width-based loss function \cite{pearce2018high} with a goal similar to LUBE, as shown in the Eq.\ref{eq:piloss}. The \textit{mean prediction interval width} (MPIW) is defined as  $| y_u - y_l |$, and the \textit{prediction interval coverage width} (PICP) measures the average probability that the PI covers the ground truth. The total loss encourages the PI to be narrow while having a higher coverage probability above the prescribed confidence level $\alpha$.
    \begin{equation}\label{eq:piloss}\footnotesize
        \text{Loss} = \text{MPIW} + \lambda* \max(0, (1-\alpha) - \text{PICP})^2.
    \end{equation}
    
    Recent approaches frame prediction interval learning as a constrained optimization problem. This optimization problem can be viewed from two perspectives: primal and dual perspectives. The primal perspective frames the objective as minimizing the PI width under the constraint that the PI attains a coverage probability larger than the confidence level \cite{chen2021learning}, which is expressed as follows:
    \begin{equation}\footnotesize
    \min_{L, U\in \mathcal{H}, L<U} \mathbb{E}_{\boldsymbol{x}\sim \pi(\boldsymbol{x})}(U(\boldsymbol{x}) - L(\boldsymbol{x})) \ s.t. \  p_{\pi}(y\in [L(\boldsymbol{x}), U(\boldsymbol{x})])>1-\alpha.
    \end{equation}
    where $\mathbb{E}_{\boldsymbol{x}\sim \pi(\boldsymbol{x})}$ denotes the expectation concerning the marginal distribution of  input samples $\boldsymbol{x}$, and $p_{\pi} $ denote the probability of the input-output pair distribution. To enforce the optimality and feasibility of the optimization problem, the tradeoff is developed through the studying of two characteristics of this approach: Lipschitz continuous model class \cite{virmaux2018lipschitz} and Vapnik–Chervonenkis (VC)-subgraph class \cite{chen2021learning}.  On the other hand, the dual perspective frames the objective as maximizing PI coverage probability subject to a fixed global budget constraint on average PI width in a batch setting \cite{rosenfeld2018discriminative}.     Researchers presented a discriminative learning framework that optimizes the expected error rate under a budget constraint on the interval sizes. This approach avoids single-point loss and provides a statistical guarantee of generalization for the entire population.  In contrast to the primal setup,  the dual perspective in batch learning constructs the prediction interval of a group of test points simultaneously, reducing the computational overhead \cite{rosenfeld2018discriminative}.

     \begin{equation}\footnotesize
    \min_{f\in \mathcal{F}} \mathbb{E}_{(\boldsymbol{x},y)\sim \pi(\boldsymbol{x},y)}L(y, U(\boldsymbol{x}), L(\boldsymbol{x})) \ s.t. \  
    \sum_i(U(\boldsymbol{x}_i)- L(\boldsymbol{x}_i)) < B.
    \end{equation}



\subsubsection{Deep Generative Model}

Deep generative models (DGMs) are a family of probabilistic models that aim to learn the complex, high-dimensional data distribution $p_{\text{data}}(\boldsymbol{x})$  with DNN. DGMs are capable of learning the intractable data distribution in the high-dimensional feature space $\mathcal{X} \subseteq \mathbb{R}^{n}$ from a large number of independent and identically distributed observed samples $\{\boldsymbol{x}_i\}_{i=1}^{m}$. Specifically, they learn a probabilistic mapping from some latent variables $\boldsymbol{z} \in \mathbb{R}^d$ that follow a tractable distribution, such as $\mathcal{N}(\mathbf{0},\mathbf{I})$ to the data distribution $p_{\text{data}}(\boldsymbol{x})$. Mathematically, the generative model can be defined as the mapping function $g_{\boldsymbol{\theta}}(\cdot): \mathbb{R}^d \rightarrow\mathbb{R}^n$, where $d$ and $n$ are the dimensions of latent variable and original data, respectively. A deep generative model is capable of capturing  probabilistic distribution for high-dimensional structured outputs (e.g., images). 

{\color{black}
The basic idea is to employ a DGM to learn the \textit{predictive distribution} $p(\boldsymbol{y}|\boldsymbol{x})$  given the supervised training data pairs $\{ (\boldsymbol{x}_i, \boldsymbol{y}_i) \}_{i=1}^{m}$. In this subsection, we use bold $\boldsymbol{y}_i$ to denote the high-dimensional structured outputs (e.g., images). It should be noted that to learn the \textit{predictive distribution} instead of the data distribution in feature space, the \textit{conditional deep generative model} (cDGM) \cite{sohn2015learning} should be employed. Generally speaking, cDGM-based uncertainty quantification models learn a conditional density over the prediction $\boldsymbol{y}$, given the input feature $\boldsymbol{x}$. This amounts to learning a model $g_{\boldsymbol{\theta}}(\boldsymbol{z}, \cdot): \mathbb{X} \rightarrow \mathbb{Y}$ such that the generative model $g(\boldsymbol{z},\boldsymbol{x})$ with $\boldsymbol{z}\sim p(\boldsymbol{z})$ approximates the true unknown distribution $p_{\text{true}}(\boldsymbol{y}|\boldsymbol{x})$. The variability of the prediction distribution $p(\boldsymbol{y}|\boldsymbol{x})$ is encoded into the latent variable $\boldsymbol{z}$ and the generative model. During inference, for any $\boldsymbol{x}\in \mathbb{X}$, we can generate $m$ samples of $\boldsymbol{y}_j$ with $g_{\boldsymbol{\theta}}(\boldsymbol{z}_j,\boldsymbol{x})$ and $\boldsymbol{z}_j\sim p(\boldsymbol{z})$. By analyzing the variability of the samples $\{\boldsymbol{y}_j\}_{j=1}^m$, 
 we can quantify prediction uncertainty. 

In the following subsection, we examine three types of deep generative models: the variational autoencoder (VAE) \cite{kingma2013auto}, the generative adversarial network (GAN) \cite{goodfellow2020generative}, and the diffusion model \cite{ho2020denoising}. The VAE, a likelihood-based generative model, is trained by maximizing the evidence lower bound (ELBO) of the likelihood function. GANs, on the other hand, are implicit generative models trained through a two-player zero-sum game framework. Lastly, the diffusion model is a probabilistic generative framework that employs a multi-step denoising process. We explore how each of these frameworks can be applied to estimate prediction uncertainty.
}

    \textbf{VAE-based model}: The VAE model consists of two modules: an \textit{encoder} and a \text{decoder}. The encoder network $q_{\phi}(\boldsymbol{z}|\boldsymbol{x})$ aims to embed the high dimensional structural output $\boldsymbol{x}$ into a low-dimensional code $\boldsymbol{z}$, that captures the inherent ambiguity or noise of the input data. The decoder $p_{\boldsymbol{\theta}}(\boldsymbol{x}|\boldsymbol{z})$ aims to reconstruct the input feature.  
     VAE model has been popular for modeling structured output uncertainty, especially for tasks on image data, because of its capability to model global and local structure dependency in regular grid images. Specifically, two kinds of frameworks based on the VAE model have been proposed to account for the data uncertainty arising from \textit{input noise} and \textit{target output noise}:  The first category aims to capture noise present in the input samples.  The basic idea is to embed each sample as a Gaussian distribution instead of a deterministic embedding in the low dimensional latent space, where the mean represents the feature embedding and variance represents the uncertainty of the embedding \cite{chang2020data}. The method accounts for varying noise levels inherent in the dataset, which is ubiquitous in many kinds of real-world datasets, for example, face image recognition \cite{chang2020data}, medical image reconstruction \cite{edupuganti2020uncertainty}. 
     This probabilistic embedding framework leverages the VAE architecture to estimate the embedding and uncertainty simultaneously.  
     The second category aims to capture the noise that lies in the target outputs, where the ground truth is imperfect, ambiguous, or corrupted. This scenario is common in the medical domain \cite{lee2020structure}, where the objects in the image are ambiguous and the experts may not reach a consensus on the class of the objects (large uncertainty). Thus for segmentation or classification tasks, the model should be aware of the prediction uncertainty.   To capture the prediction uncertainty in the target outputs, the conditional VAE (cVAE)  \cite{sohn2015learning} framework is adopted. Specifically, cVAE formulates the prediction distribution as an integration over the latent embedding $\boldsymbol{z}$, 
     \begin{equation}\footnotesize
         p(\boldsymbol{y}|\boldsymbol{x}) = \int p(\boldsymbol{y}|\boldsymbol{x}, \boldsymbol{z} )p(\boldsymbol{z}|\boldsymbol{x})d\boldsymbol{z} \approx \frac{1}{n}\sum_{j=1}^n p(\boldsymbol{y}|\boldsymbol{x}, \boldsymbol{z}_j ),  \ \text{where} \ z_j \sim p(\boldsymbol{z}).
     \end{equation}
     The cVAE model is trained by maximizing the evidence lower bound of the likelihood. Then, during inference, multiple latent features $\boldsymbol{z}_j$ can be drawn from the prior distribution, and the integration over latent $\boldsymbol{z}$ can be approximated with the sampling distribution \cite{prokudin2018deep}. 
     Probabilistic U-Net model  \cite{kohl2018probabilistic} combines the architecture of cVAE and U-Net model by treating the U-Net model as the encoder to produce a probabilistic segmentation map. The U-Net model can capture multi-scale feature representations in the low-dimensional latent space to encode the potential variability in the segmentation map. These models are illustrated in Fig.~\ref{fig:vae} (a) and (b).

In summary, the VAE-based framework can take into consideration the data uncertainty coming from the input noise or the target output noise and can integrate state-of-the-art neural network architectures into its framework, making it more flexible for many kinds of applications. The key success lies in modeling the joint probability of all samples (pixels) in the image.  The approach is suitable for structured uncertainty quantification (e.g., image grid structure, graph structure) by learning the implicit joint distribution of the structure. 


\begin{figure}
    \centering
    \subfloat[Input uncertainty.]{\includegraphics[height=1.1in]{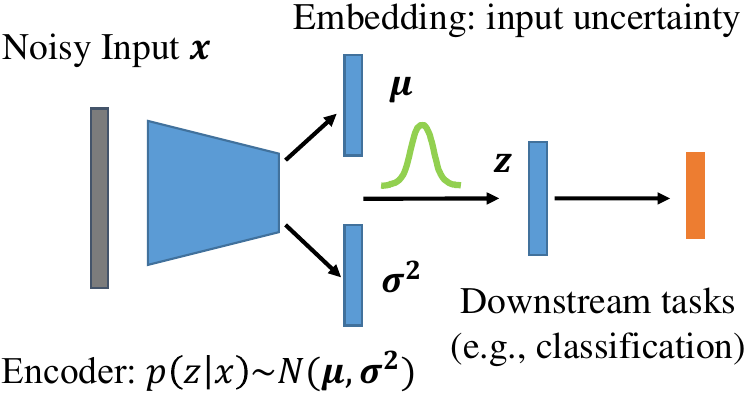}} 
    \hspace{0.2in}
   \subfloat[Target prediction uncertainty.]{\includegraphics[height=1.1in]{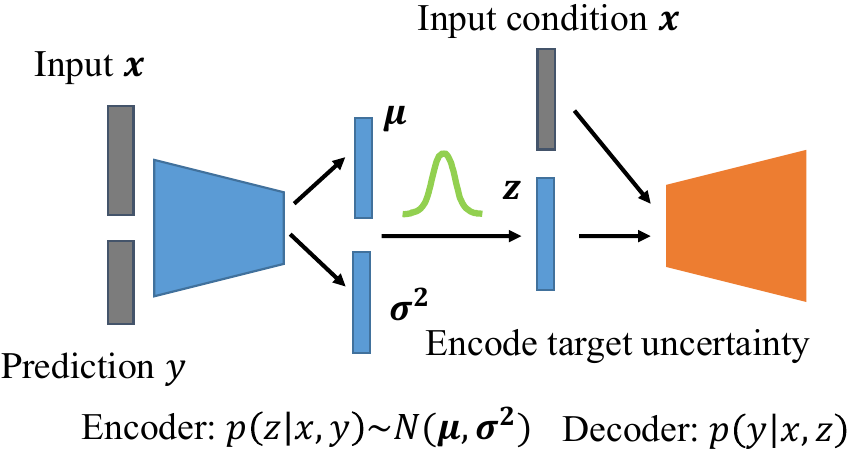}}
    \caption{VAE framework for uncertainty quantification.}
    \label{fig:vae}
\end{figure}





    \textbf{GAN-based generative model}: GAN is a type of generative model trained with a two-player zero-game. It consists of a \textit{generator} and a \textit{discriminator}. In conditional GAN (cGAN), the generator takes the input $x$ and random noise $\boldsymbol{z}$ as input and generates the target variables $y$: $\mathcal{G}: (\boldsymbol{x},\boldsymbol{z}) \rightarrow \boldsymbol{y} $. The discriminator is trained to distinguish between generated samples and ground-truth samples.  GAN has been adopted in many domains. For example, in the transportation domain, GAN has been used for traffic volume prediction \cite{mo2022trafficflowgan}. The flow model is integrated with GAN to enable likelihood estimation and better uncertainty quantification. Another approach  \cite{oberdiek2022uqgan,gao2022wasserstein}  extends this method using the Wasserstein GAN \cite{arjovsky2017wasserstein} with gradient penalty to improve model convergence. 
    The key advantage of deep generative modeling for uncertainty quantification is that it directly parameterizes a deep neural network to represent the prediction distribution without needing an explicit distribution format. Moreover, it can integrate a physics-informed neural network for better uncertainty estimation of physical science \cite{daw2021pid}. 
    However, GAN-based models are harder to train, especially for GAN-based models. Model convergence is not guaranteed. 

    \textbf{Diffusion-based generative model}: {\color{black} 
    Diffusion models are a family of probabilistic generative models that progressively destroy data by injecting noise in the forward diffusion process,
then learn to reverse this process to generate new data samples through a backward process \cite{yang2023diffusion}. The forward diffusion process gradually adds noise to the data 
according to a variance schedule $\beta_t$: $q(\mathbf{x}_t | \mathbf{x}_{t-1}) = \mathcal{N}(\mathbf{x}_t; \sqrt{1 - \beta_t} \, \mathbf{x}_{t-1}, \beta_t \, \mathbf{I})$.  The reverse process is defined as:
$
p_\theta(\mathbf{x}_{t-1} | \mathbf{x}_t) = \mathcal{N}(\mathbf{x}_{t-1}; \mu_\theta(\mathbf{x}_t, t), \Sigma_\theta(\mathbf{x}_t, t))
$, 
where \( \mu_\theta \) and \( \Sigma_\theta \) are learned with a neural network that aims to reconstruct (denoise) \( \mathbf{x}_{t-1} \) from \( \mathbf{x}_t \) at each step. The training objective supervises the model to recover data from noise accurately. Diffusion models have been applied to images \cite{ho2020denoising}, videos \cite{bar2024lumiere}, time series \cite{tashiro2021csdi}, and scientific simulations \cite{finzi2023user}. 
Similar to conditional VAEs and GANs, conditional diffusion models can be applied to uncertainty quantification of model outputs given input features~\cite{finzi2023user,andrae2024continuous,li2024generative}. 
Latent diffusion models enhance the efficiency of diffusion models for high-resolution images or video generation by operating on lower dimensional latent embeddings (e.g., learned through an autoencoder) \cite{rombach2022high,neumeier2024reliable,gao2024bayesian}. As summarized in Table~\ref{tab:datauq} below, the primary advantages of diffusion models are their expressiveness to model complex high-dimensional distributions and generate high-quality samples \cite{yang2023diffusion}. A major drawback, however, is the slow training and inference due to the iterations of hundreds to thousands of steps \cite{yang2023diffusion}. There is ongoing research that addresses this issue \cite{kimfast}.

{\bf Summary on Data Uncertainty:} Table~\ref{tab:datauq} compares the pros and cons of existing approaches for data uncertainty quantification. Discriminative models are simple but unsuited for capturing structured output uncertainty. Generative can better quantify structured output uncertainty but requires multiple sampling of model outputs. Among deep generative models, diffusion models are currently more popular, but more efforts are needed to improve their efficiency.

{\color{black}
The methods for data uncertainty, including deep discriminative models and deep generative models, can also quantify total uncertainty with minimal modifications. For instance, deep generative models inherently account for data uncertainty by marginalizing over latent variables. Model uncertainty can be estimated by computing the variance of predictions generated from multiple latent samples, similar to Bayesian neural networks. For deep discriminative models, model uncertainty can be assessed by measuring the variance of predictions from multiple parameterized models. A more systematic review of techniques for disentangling and quantifying multiple types of uncertainty is in Section 4.3.
}
    }



\begin{table*}[ht]
\footnotesize
\centering
\caption{A comparison of UQ methods for data uncertainty}
\label{tab:datauq}
{\color{black}
\begin{tabular}{|l|l|l|l|}
\hline
{\bf Model} &
{\bf  Approach} &
{\bf  Pros} &
 {\bf  Cons} \\ \hline
\multirow{2}{*}{\begin{tabular}[c]{@{}l@{}}{\bf Deep} {\bf Discriminative model}\\ {\bf Pros}: No need to modify \\ network architecture.\\ {\bf Cons}: Not suitable for  structured output.\end{tabular}} &
  \begin{tabular}[c]{@{}l@{}}  Predict a parametric\\ distribution \cite{kendall2017uncertainties,guillaumes2017mixture}.  \end{tabular} &
  \begin{tabular}[c]{@{}l@{}}Simple model training.\end{tabular} &
  \begin{tabular}[c]{@{}l@{}}Assume a parametric \\ output distribution.\end{tabular} \\ \cline{2-4} 
 &
  \begin{tabular}[c]{@{}l@{}}  Non-parametric: \\ predict an interval \\  \cite{pearce2018high,wu2021quantifying,rosenfeld2018discriminative,chen2021learning}.  \end{tabular} &
  \begin{tabular}[c]{@{}l@{}}No rigid assumption on\\ output distribution.\end{tabular} &
  \begin{tabular}[c]{@{}l@{}}Need to design new\\ training loss.\end{tabular} \\ \hline
\multirow{2}{*}{\begin{tabular}[c]{@{}l@{}}{\bf Deep Generative } {\bf  model}\\{\bf Pros}: Capture   uncertainty \\ for structured output data.\\{\bf Cons}: Require  multiple sampling \\for uncertainty quantification.  \end{tabular}} &\begin{tabular}[c]{@{}l@{}} 
  VAE-based model \\  \cite{chang2020data,edupuganti2020uncertainty,prokudin2018deep,kohl2018probabilistic}. \end{tabular}  &
  \begin{tabular}[c]{@{}l@{}}Stability  in training,  \\ probabilistic outputs. \\ \end{tabular} &
  \begin{tabular}[c]{@{}l@{}}Less expressive  \\ compared with GAN \\ and diffusion models.\end{tabular} \\ \cline{2-4} 
 &\begin{tabular}[c]{@{}l@{}}
  GAN-based model \\  \cite{oberdiek2022uqgan,gao2022wasserstein,mo2022trafficflowgan}. \end{tabular}&
  \begin{tabular}[c]{@{}l@{}} More expressive \\ than VAE.\end{tabular} &
  \begin{tabular}[c]{@{}l@{}}Instability in training \\ and lack of probabilistic \\ framework.\end{tabular} \\ 
  \cline{2-4} 
 &\begin{tabular}[c]{@{}l@{}}
  Diffusion-based model \\  \cite{shu2024zero,du2023diffusion,berryshedding,finzi2023user}. \end{tabular}&
  \begin{tabular}[c]{@{}l@{}} More expressive than VAE, \\ probabilistic framework.\end{tabular} &
  \begin{tabular}[c]{@{}l@{}} High computational \\ cost. \end{tabular} \\ \hline
\end{tabular}}
\end{table*}

\subsection{Model and data uncertainty}
Besides considering the data and model uncertainty separately, many frameworks attempt to jointly consider the two kinds of uncertainty for more accurate quantification. In this part, we will review existing frameworks that aims to quantify both types of uncertainty simultaneously.  

\subsubsection{Approaches combining data and model uncertainty}\label{subsec:combinedatamodel}
A straightforward way to consider both data and model uncertainty is to select one of the approaches in each category and combine them in a single framework. Below, we will introduce some major ways to combine approaches for data and model uncertainty and their potential drawbacks.

     \textbf{Combine BNN model with prediction distribution}: The method aims to capture both data and model uncertainty within a single framework   \cite{kendall2017uncertainties} by combining  BNN with prediction distribution. The model uncertainty is captured with the BNN approximation approach.  Specifically, MC drop-out is adopted due to its simplicity for implementation. For each dropout forward pass, a sample of the weights is drawn from the weight distribution approximation  $\mathbf{W}_t\sim \text{Bernoulli}(p)$, where $p$ is the dropout rate, then one forward prediction can be made with the weight by $y_t = p(y|\boldsymbol{x}, \mathbf{W}_t)$.
To obtain the data uncertainty, the output is formulated as a parameterized Gaussian distribution instead of point estimation $[y_t, \sigma_t^2] = p(y|\boldsymbol{x}; \mathbf{W}_t)$, where $y_t$ is the target variable mean prediction and $\sigma_t^2$ is the prediction variance for a single forward prediction. With multiple dropout forward passes, we have a set of $T$ prediction samples $\{y_t, \sigma^2_t \}_{t=1}^{T}$. The predictive uncertainty in the combined model can be approximated with the law of total variance expressed as $\text{Var}(y)$ in Eq.~\ref{eq:totalvar}. The intuition behind this equation is that total uncertainty comprises two parts, the last $\frac{1}{T}\sum_{t=1}^T\sigma_t^2$ represents data uncertainty on average, and the first part $\frac{1}{T}\sum_{t=1}^T y_t^2 - (\frac{1}{T}\sum_{t=1}^T y_t)^2$ represents the disagreement across  $T$ MC-dropout models, which captures  model uncertainty. 
\begin{equation}\label{eq:totalvar}\footnotesize
    \text{Var}(y) \approx \frac{1}{T}\sum_{t=1}^T y_t^2 - (\frac{1}{T}\sum_{t=1}^T y_t)^2 + \frac{1}{T}\sum_{t=1}^T\sigma_t^2.
\end{equation}

     \textbf{Combine ensemble model with prediction distribution}: This approach \cite{lakshminarayanan2017simple} combines the ensemble method with prediction distribution. The deep ensemble method constructs an ensemble of DNN models $\mathcal{M} = \{\mathcal{M}_i\}_{i=1}^K$, where each model $\mathcal{M}_i$ can be set with different parameters, or architecture choices. The model uncertainty is expressed as the variance or "disagreement" among the ensemble models.  In this way, the output of each model is modified as a parameterized distribution to capture the data uncertainty. Similar to MC-dropout, we have an ensemble of prediction distribution $\{p(y|\boldsymbol{x}, \mathcal{M}_i)\}_{i=1}^K$. In this part, we take the classification problem as an example, where the prediction distribution is a parameterized categorical distribution. The total uncertainty is captured with the entropy of average prediction distribution $\mathcal{H}(\mathbb{E}_{p(\mathcal{M}_i)}p(y|\boldsymbol{x},\mathcal{M}_i))$, and the data uncertainty is the average entropy of each model, expressed as $\mathbb{E}_{p(\mathcal{M}_i)}\mathcal{H}(p(y|\boldsymbol{x},\mathcal{M}_i)) $. The model uncertainty can be expressed with the mutual information between the prediction and the ensemble model $y, \mathcal{M}$ as expressed in Eq.~\ref{eq:ensemblewithdistr}.
     \begin{equation}\label{eq:ensemblewithdistr}\footnotesize
         \text{MI}(y,\mathcal{M}) = \mathcal{H}(\mathbb{E}_{p(\mathcal{M}_i)}p(y|\boldsymbol{x},\mathcal{M}_i)) - \mathbb{E}_{p(\mathcal{M}_i)}\mathcal{H}(p(y|\boldsymbol{x},\mathcal{M}_i)).
     \end{equation}
     
     \textbf{Combine ensemble model with prediction interval}: Since the prediction interval constructed in some approaches accounts only for the data noise variance, not the model uncertainty. To improve the total uncertainty estimation, ensemble methods are adopted to combine prediction intervals with ensemble methods to account for model uncertainty arising from model architecture misspecification, parameter initialization, etc. \cite{pearce2018high}. Specifically, Given an ensemble of models trained with different model specifications or sub-sampling of training datasets, where the model prediction intervals are denoted as $[y^{ij}_l, y^{ij}_u]$ for sample $i=\{1,...,n\}$ and model $j=\{1,...,m\}$, the model uncertainty can be captured by the variance of the lower bound $\sigma_{l}^{(i)^2}$ and upper bound variance $\sigma_{u}^{(i)^2}$.  For example, the uncertainty of the lower bound is given by:
     \begin{equation}\footnotesize
     \begin{split}
    &\sigma_{l}^{(i)^2} = \frac{1}{m-1}\sum_{j=1}^m(y_l^{(ij)} - \hat{y}_l^{(i)})^2, \ \text{where} \ \hat{y}_l^{(i)} = \frac{1}{m}\sum_{j=1}^my_l^{(ij)}, \\
        &\sigma_{u}^{(i)^2} = \frac{1}{m-1}\sum_{j=1}^m(y_u^{(ij)} - \hat{y}_u^{(i)})^2, \ \text{where} \ \hat{y}_u^{(i)} = \frac{1}{m}\sum_{j=1}^my_u^{(ij)}.
     \end{split}
     \end{equation}
     Then the new prediction interval $[\tilde{y}_l,\tilde{y}_u]$ with $95\%$ confidence level can be constructed as: 
     
          \begin{equation}\footnotesize
     \begin{split}
    &\tilde{y}_l = y_l - 1.96\sigma_{l}^{(i)^2}, \  \text{and} \ 
        \tilde{y}_u = y_u + 1.96\sigma_{u}^{(i)^2}.
     \end{split}
     \end{equation}
     The constructed interval can reflect both the data uncertainty and model uncertainty. However, this approach relies on variance from the ensemble's lower and upper bounds, which lacks theoretical justification due to the independent treatment of the two boundaries. To overcome this limitation, one recent approach proposes a split normal aggregation method to aggregate the prediction interval ensembles into final intervals \cite{salem2020prediction}. Specifically, the method fits a split normal distribution (two pieces of normal distribution) over each prediction interval, and then the final prediction will become a mixture of split normal distribution. The PI can be derived from the $1-\alpha$ quantile of the cumulative distribution.

 In summary, to capture both the data and model uncertainty, existing literature can combine the methodologies in the two categories. There are several limitations to the combination approaches: first, the BNN or ensemble models require multiple forward passes for the prediction, which introduces computation overhead and extra storage. Efficiency is a concern. Second, the simple combination of data and model uncertainty lacks a theoretical guarantee, which requires post hoc calibration of the model.


\subsubsection{Evidential deep learning}

To overcome the computational challenge for the combination approaches, evidential deep learning was proposed to use one single deterministic model to capture both the data and model uncertainty without multiple forward passes of the neural network \cite{malinin2018predictive,charpentier2020posterior,sensoy2018evidential,bao2021evidential}. The intuition of evidential deep learning is to predict class-wise evidence instead of directly predicting class probabilities.  In the following section, we review these methodologies, including their advantages and disadvantages. 

As discussed in the aforementioned sections, for classification problems, existing
deep learning \begin{wrapfigure}{r}{7.5cm}  
    \centering
    {\includegraphics[height=1.0in]{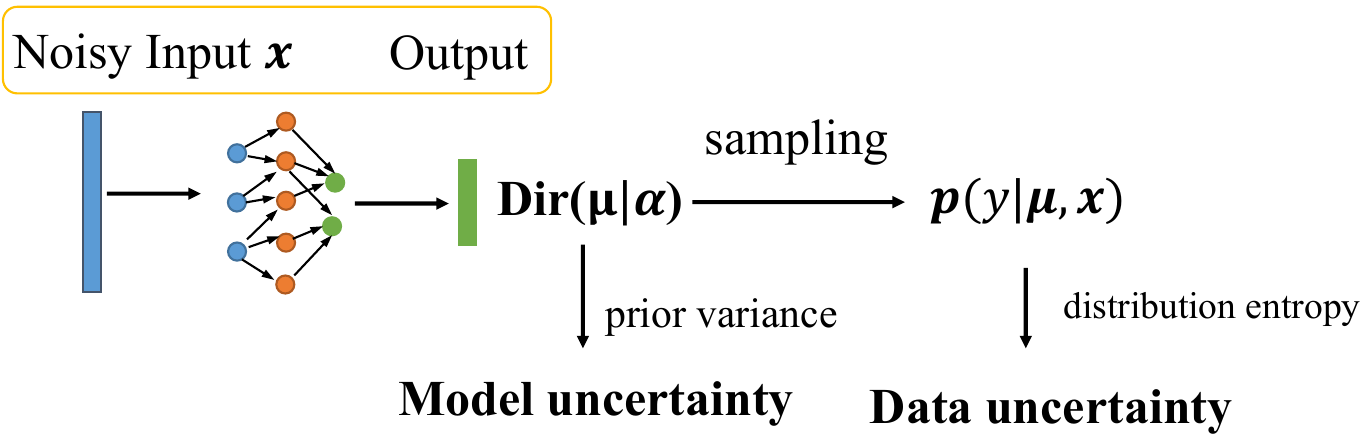}}
    \caption{Evidential deep learning architecture.}
    \label{fig:evidential}
\end{wrapfigure}  based models explicitly or implicitly predict class probabilities (categorical distribution parameters)  with softmax-layer parameterized by DNNs to quantify prediction uncertainty. However, softmax prediction uncertainty often tends to be overconfident \cite{hendrycks2016baseline}. Evidential deep learning is developed to overcome the limitation by introducing evidence theory \cite{audun2018subjective} to neural network frameworks. 
The goal of evidential deep learning is to construct predictions based on evidence and predict the parameters of Dirichlet density. For example, considering the 3-class classification problem, a vanilla neural network directly predicts the categorical distribution for each class $\pi = {\pi_1, \pi_2, \pi_3}$ with $\sum_i \pi_i = 1$. However, this approach can only represent a point estimation of prediction distribution. On the other hand, evidential deep learning aims to predict the \textit{evidence} for each class
$\boldsymbol{\alpha} = \{\alpha_1,\alpha_2, \alpha_3 \}$ with the constraint $\boldsymbol{\alpha} > 0 $, which can be considered as the parameters of Dirichlet distribution \cite{sensoy2018evidential}. The framework is shown in Fig~\ref{fig:evidential}, where the output is the \textit{evidence} $\boldsymbol{\alpha}$ for each class, and the prediction distribution is sampled from the Dirichlet distribution. The expected prediction distribution for each class is $p_i = \frac{\alpha_i}{\sum_{c=1}^3\alpha_c}$, whose entropy represents data uncertainty. On the other hand,  model uncertainty is reflected by the total evidence $\sum_i\alpha_i$, which means the more evidence we collect, the more confident the model is. 

Mathematically, evidential deep learning aims to use a neural network to learn the prior distribution of categorical distribution parameters, which is represented by the Dirichlet distribution.  The Dirichlet distribution is parameterized by its concentration parameters $\boldsymbol{\alpha}$ (evidence) where $\alpha_0$ is the sum of all $\alpha_i$ and is referred to as the precision of the Dirichlet distribution. A higher value of $\alpha_0$ will lead to sharper distribution and lower model uncertainty. As shown in Eq.\ref{eq:dir} below, the $\text{Dir}(\boldsymbol{\mu}|\boldsymbol{\alpha})$ defines a probability density function over the k-dimensional random variable $\boldsymbol{\mu} = [{\mu_1},..., {\mu}_k]$, where $k$ is the number of classes, $\boldsymbol{\mu}$ belongs to the standard $k-1$ simplex (${\mu_1}+...+ {\mu}_k = 1$ and $ {\mu}_i \in [0,1]$ for all $i \in {1,..., k}$), and can be regarded as the categorical distribution parameters. 
The relationship between the Dirichlet distribution and uncertainty quantification can be illustrated using a 2-simplex in Fig.~\ref{fig:data_uq}. The random variable $\boldsymbol{\mu} = [{\mu}_1, {\mu}_2, {\mu}_3]$ is represented by its Barycentric coordinates on the 2-simplex. The Barycentric coordinate is a coordinate system where points are located inside a simplex, and the value in each coordinate can be interpreted as the fraction of mass placed at each corresponding vertex of the simplex. Fig.~\ref{fig:data_uq} (a) shows a scenario where the evidence parameters are equal, resulting in indistinguishable classes and implying high data uncertainty (high entropy for the sampled $\boldsymbol{\mu}$). As the total evidence $\sum_i \alpha_i$ increases, the density becomes more concentrated, which means the model uncertainty decreases, while the data uncertainty remains fixed. Fig.~\ref{fig:data_uq} (b) shows a scenario with fixed model uncertainty, as the sum of evidence parameters remains constant. When the evidence becomes imbalanced, the density becomes more concentrated toward one class, thus decreasing the data uncertainty.
\begin{equation}\label{eq:dir}\footnotesize
\text{Dir}(\boldsymbol{\mu}|\boldsymbol{\alpha}) = \frac{\mathrm{T}(\alpha_0)}{\prod_{c=1}^K\mathrm{T}(\alpha_c)}\prod_{c=1}^K{\mu}_c^{\alpha_c - 1}, \alpha_c > 0, \ \alpha_0 = \sum_{c=1}^K \alpha_c.
\end{equation}
 
\begin{figure}
    \centering
   \subfloat[Keep data uncertainty fixed , and model uncertainty decrease from left to right.]{ \includegraphics[height=0.8in]{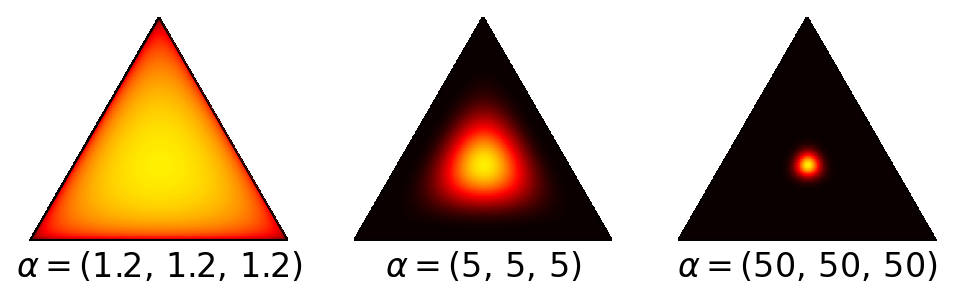}} \hspace{3mm}
   \subfloat[Keep the model uncertainty fixed, and  the data uncertainty decreases from left to right (the entropy of the categoric distribution decreases).]{\includegraphics[height=0.8in]{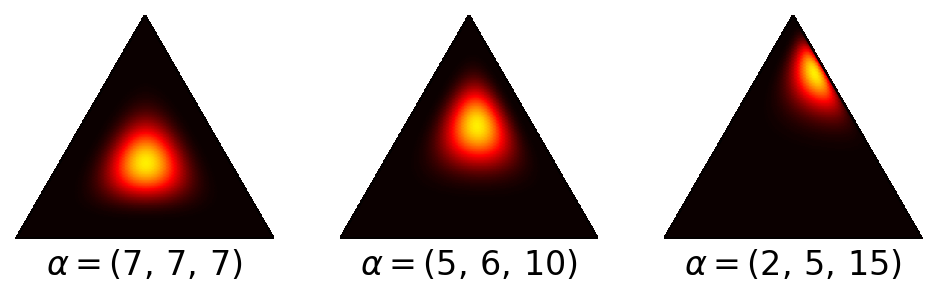}} 
    \caption{Dirichlet distribution density visualization.}
    \label{fig:data_uq}
\end{figure}

Due to the intriguing property of Dirichlet distribution, evidential deep learning directly predicts the parameters of Dirichlet density. For example, 
the Dirichlet prior network (DPN) \cite{malinin2018predictive} learns the concentration parameter $\boldsymbol{\alpha}$ for the Dirichlet distribution $\boldsymbol{\alpha} = \boldsymbol{f}(\boldsymbol{x}, \boldsymbol{\theta})$. The categorical distribution parameters are then drawn from the Dirichlet distribution as $p(\boldsymbol{\mu}|\boldsymbol{x}, \boldsymbol{\theta} ) = \text{Dir}(\boldsymbol{\mu}|\boldsymbol{\alpha})$. 
The predicted class probability is the average over possible values of  $\boldsymbol{\mu}$.
\begin{equation}\footnotesize
    p(w_c|\boldsymbol{x}, \boldsymbol{\theta}) = \int p(w_c|\boldsymbol{\mu})p(\boldsymbol{\mu}|\boldsymbol{x}, \boldsymbol{\theta} ) = \frac{\alpha_c}{\alpha_0}.
\end{equation}

To measure uncertainty from the Dirichlet distribution, the total uncertainty is computed as the entropy of the average predictive distribution, while data uncertainty is determined by averaging the entropy across each realization of $\boldsymbol{\mu}$. Several approaches have extended the prior network to handle regression tasks, and a posterior network has been introduced to enhance the reliability of uncertainty estimation. Additionally, other second-order methods have been proposed to jointly quantify data and model uncertainty within a single deterministic framework \cite{hofman2024quantifying,kotelevskii2024predictive,bengs2023second}. 

The advantage of evidential deep learning is the approach  requires only a single forward pass during inference and is much more computationally efficient. The approach also explicitly distinguishes data and model uncertainty in a principled way. The disadvantage is that the training stage is more complex and does not guarantee the same prediction accuracy as a vanilla network, and it does not leverage many advances in modern DNNs (e.g., large scale pretraining \cite{brown2020language} and sampling-based architectures \cite{yang2023diffusion}). Furthermore, the training process typically requires OOD samples to learn effective representations, which increases the required dataset size. 


\begin{table*}[h]
\footnotesize
\centering
\caption{A comparison of UQ methods for both model and data uncertainty}
\label{tab:combineuqtable}
{\color{black}
\begin{tabular}{|l|l|l|l|}
\hline
{\bf Model} & 
{\bf   Approaches } &
{\bf   Pros} &
{\bf   Cons} \\ \hline
\multirow{3}{*}{\begin{tabular}[c]{@{}l@{}}{\bf Combination of} \\ {\bf existing approaches}\end{tabular}} &
  \begin{tabular}[c]{@{}l@{}}Combine BNN \\ with  prediction \\distribution \cite{kendall2017uncertainties}.\end{tabular} &
  \begin{tabular}[c]{@{}l@{}}Capture uncertainty \\ from   both data noise \\ and model parameters.\end{tabular} &
  \begin{tabular}[c]{@{}l@{}}Require multiple  \\  forward pass \\ during inference.\end{tabular} \\ \cline{2-4} 
 &
  \begin{tabular}[c]{@{}l@{}}Combine ensemble \\ with   prediction \\ distribution \cite{lakshminarayanan2017simple}. \end{tabular} &
  \begin{tabular}[c]{@{}l@{}}Capture uncertainty \\ from  both data  noise \\ and mode  architectures.\end{tabular} &
  \begin{tabular}[c]{@{}l@{}}Require more \\ computation and \\ storage requirement.\end{tabular} \\ \cline{2-4} 
 &
  \begin{tabular}[c]{@{}l@{}}Combine ensemble \\ with   prediction \\ interval \cite{pearce2018high,salem2020prediction}.\end{tabular} &
  \begin{tabular}[c]{@{}l@{}}Capture both data and  \\ model uncertainty and \\ do not need explicit \\ parametric distribution  form.\end{tabular} &
  \begin{tabular}[c]{@{}l@{}}Require modification \\  on existing DNN \\ training process.\end{tabular} \\ \hline
\begin{tabular}[c]{@{}l@{}}{\bf Evidential deep}\\ {\bf learning}\end{tabular} &
  \begin{tabular}[c]{@{}l@{}}Evidential deep \\ learning \cite{malinin2018predictive,charpentier2020posterior,sensoy2018evidential,bao2021evidential}.\end{tabular} &
 {\begin{tabular}[c]{@{}l@{}}
  Computationally efficient \\ relative to combination \\approaches.\end{tabular}} &
 { \begin{tabular}[c]{@{}l@{}} Strong assumption on \\ prior distribution. \\ Difficult to train. \end{tabular}} \\ \hline
\begin{tabular}[c]{@{}l@{}}{\bf Conformal}\\ {\bf Prediction}\end{tabular} &
  \begin{tabular}[c]{@{}l@{}}Prediction  set \\ or interval \cite{lu2022fair,caprio2023imprecise,hechtlinger2018cautious,karimi2023quantifying}.\end{tabular} &
 {\begin{tabular}[c]{@{}l@{}} No rigid assumption\\ in
  distribution. \end{tabular}} &
 { \begin{tabular}[c]{@{}l@{}}Require exchangeability \\ on sample distribution.  \end{tabular} }\\ \hline
\end{tabular}}
\end{table*}

\subsubsection{Conformal Prediction}

Though existing UQ approaches can  quantify total uncertainty from various sources, a major limitation is the lack of formal guarantees. To address this issue, conformal prediction (CP) is  a post-processing approach that constructs a finite prediction set with a statistical guarantee to cover the true label.
Conformal prediction, which belongs to distribution-free uncertainty quantification is model-agnostic, and provides formal results for marginal coverage, defined as the average probability that the true class is contained in the prediction set \cite{angelopoulos2023conformal}. Formally, this coverage guarantee states:
\begin{equation}\label{eq:conformalclass}
     \mathbb{P}(y\in \mathcal{C}(x)) \ge 1 - \alpha.
\end{equation}
where $\boldsymbol{x}$ is a sample instance, and $y$ is the ground truth. Unlike traditional confidence intervals which typically assume a specific distribution of data or residuals, conformal prediction provides a non-parametric and distribution-free approach \cite{karimi2023quantifying}. This is advantageous for deep learning, where traditional assumptions are often violated due to model complexity and non-linearity. 

{\color{black}
There are several approaches to developing conformal predictions for deep learning models. One approach is to apply conformal methods directly to pre-trained deep learning models \cite{lu2022fair}. Second, credal Bayesian deep learning trains an (uncountable) infinite ensemble of BNNs using only finitely many elements and outputs a prediction set for total uncertainty estimation \cite{caprio2023imprecise}. Third, density-based deep conformal prediction models uncertainty from out-of-distribution samples by considering distances between samples explicitly \cite{hechtlinger2018cautious}. Common conformal prediction methods assume that data instances are independently and identically distributed (i.i.d.). Recent efforts extend conformal prediction beyond this assumption to graph data by introducing a permutation invariance condition \cite{zargarbashi2023conformal,xu2023conformal} and to time series data through adaptive conformal prediction \cite{zaffran2022adaptive}.

{\bf Summary on UQ for Both Model and Data Uncertainty:} Table~\ref{tab:combineuqtable} compares the pros and cons of existing approaches
for both data and model uncertainty. The combination of data and model uncertainty estimation methods is simple but computationally expensive. In contrast, evidential deep learning and conformal prediction are more efficient but require more assumptions on the data distributions. 

}

{\color{black}  
\subsection{Evaluation Metrics}

The previous subsections discuss various approaches for model and data uncertainty quantification. In this part, we discuss how to evaluate the performance of an uncertainty quantification model. There are several metrics to evaluate the performance of uncertainty quantification from different perspectives. First, one method is to evaluate the calibration between the uncertainty measure and the fraction of correct prediction, called expected calibration error (ECE \cite{naeini2015obtaining}). It first discretizes the uncertainty value into a fixed number of bins and assigns each predicted probability to the bin that encompasses it. The calibration error is the difference between the fraction of correct predictions in the bin (accuracy) and the mean of the probabilities in the bin (confidence).
Given the groundtruth $\hat{y}$, the prediction from a model $Y$, and prediction uncertainty $\hat{p}$, the model is calibrated when the probability of being correct $\mathbb{P}(Y=\hat{y})$ is equal to the uncertainty measure $\hat{p}$:

\begin{equation}\label{eq:calibration}
    \mathbb{P}(Y=\hat{y}|\hat{p}) = \hat{p} \ \text{for any} \ \hat{p}.
\end{equation}

Second, to address any distortion
of results introduced by the binning procedure, the smooth ECE (smECE \cite{blasiok2023smooth}), which avoids the binning altogether by
smoothing observations using an RBF kernel. Third, the area under the ROC (AUROC) \cite{ovadia2019can} treats the problem as a binary misclassification by using uncertainty to predict the correctness of the prediction and aggregating the results over all possible decision thresholds. The intuition of this measure is that more uncertainty should indicate a lower probability of the correctness of the answers. Thus, AUROC measures the rank of the uncertainty score.  Fourth, the Brier score is a strictly proper score function \cite{brier1950verification}. It measures the mean squared difference between the predicted probability and the actual outcome. Last, Adaptive Calibration Error (ACE \cite{nixon2019measuring}) computes the calibration across all predictions for multi-classes to overcome the limitation of ECE.}

\section{Uncertainty estimation in various machine learning problems}\label{sec:mlproblems}

In this section, we discuss several major ML problems where UQ can play a critical role. 

\subsection{Out-of-distribution  detection}

A fundamental assumption in deep neural networks is that the test data distribution closely resembles the training data distribution $p_{\text{train}}(\boldsymbol{x}) \approx p_{\text{test}}(\boldsymbol{x})$. However, in complex real-world scenarios, a DNN can encounter out-of-distribution (OOD) samples that differ from the training data distribution. This can lead to significant drops in prediction performance. A DNN model should be able to recognize these situations. 

 Given a training data distribution $p(\boldsymbol{x})$, OOD data includes samples that are either unlikely under the training data distribution or outside the support of $p(\boldsymbol{x})$. Accurate detection of OOD samples is important for safety-critical applications, e.g., autonomous driving  \cite{shafaei2018uncertainty}, medical image analysis  \cite{ghoshal2021estimating}. Since OOD samples lie further away from the training samples, the model may not generalize well and could produce unstable predictions. The primary uncertainty for OOD data is concerned with model uncertainty because the model trained with in-distribution may not generalize well to other domains. 

{\color{black} \textit{Existing approaches:} Existing approaches leveraging the model uncertainty framework are much more popular for OOD detection. For example, drop-out-based BNN approaches have been applied to OOD detection and improved the performance using randomized embeddings from intermediate layers of a dropout BNN \cite{nguyen2022out} and node-based BNN \cite{trinh2022tackling}. Deep ensembles are simple and well-performing on OOD detection \cite{lakshminarayanan2017simple,zaidi2021neural}. Recent advances have developed distance-aware DNN for more accurate OOD detection by imposing constraints on the feature extracting process \cite{liu2020simple, lee2018simple}.   The evidential deep learning framework has also demonstrated its capability on OOD detection in many benchmark datasets because of the explicit distinction between two types of uncertainty in the framework \cite{charpentier2020posterior}. }




\subsection{Active Learning}

Obtaining labeled data for deep learning models can be laborious and time-consuming. Active learning  \cite{ren2021survey} aims to solve the data labeling issue by learning from a small amount of data and choosing by the model what data it requires the user to label and retrain the model iteratively. The goal is to reduce the number of labeled examples needed by using a strategy to prioritize samples worth labeling. A popular strategy is to use predictive uncertainty, prioritizing instances where predictions are most uncertain.

 The key goal in active learning is to choose observations $\boldsymbol{x}$ where obtaining labels $y$ would improve learning performance. As discussed in the background, adding samples with high data uncertainty may not improve the trained model because its inherent randomness is irreducible while more samples with model uncertainty can improve the model's performance. In this regard, model uncertainty is more important for active learning \cite{nguyen2022measure}. 
The critical challenge for active learning is to distinguish between the data and model uncertainty and utilize model uncertainty for selecting new samples.

\textit{Existing approaches}: Similar to OOD detection, approaches for model uncertainty detection can be adapted for active learning by considering uncertainty coming from the parameter, model architecture, and sample density sparsity. For example, the BNN framework considers the samples that decrease the entropy of $p(\boldsymbol{\theta}|\mathcal{D}_{\text{tr}}, \{\boldsymbol{x},y\})$ the most will be the most useful \cite{depeweg2019modeling}. The deep ensemble and MC-dropout approach can also be a straightforward way for quantifying the model uncertainty in active learning \cite{hein2022comparison}. Recent approaches propose margin-based uncertainty sampling schemes and provide convergence analysis \cite{raj2022convergence}. 

\subsection{Deep Reinforcement Learning}

 Deep reinforcement learning (DRL) aims to train an agent interacting with the environment to maximize its total rewards \cite{sutton2018reinforcement}. DRL can be regarded as learning via the Markov Decision Process, defined by the tuple $\{S, A, R, P \}$, where  $S$  is the set of states (environment conditions), $A$ is the set of actions (agent), $R$ is the function mapping state-action pairs to rewards, and $P$ is the transition probability (the probability of next state after performing actions on current state). The goal of DRL is to learn the policy $\pi$ (a function mapping given state to an action) that maximizes the sum of discounted future rewards $J(\pi) = \mathbb{E}[\sum_i\gamma^iR(s_i, a_i)]$, where $\gamma$ is the discount factor on the future reward (indicating that future rewards are less significant) \cite{sutton2018reinforcement}. In Deep Q-learning, DNN models are used to learn value functions (expected rewards for each state-action pair) given an input state. 

Due to the complex agent and environment conditions and limited training states,  two types of uncertainty sources exist: data uncertainty and model uncertainty.   Data uncertainty arises from the intrinsic randomness in the interactions between the agent and environment, affecting the reward $R$, the transition functions $P$, and the next state value distribution. To characterize the data uncertainty arising from those sources, the distributional RL  \cite{bellemare2017distributional} takes a probabilistic perspective on learning the reward functions instead of approximating the expectation of the value. Thus, the approach can be used to implement risk-aware behavior in the agent. A similar approach is proposed to quantify the data uncertainty in DRL, aiming for curiosity-based learning in the face of unpredictable transitions \cite{mavor2022stay}.  The following work \cite{dabney2018distributional} extends the parametric distribution to non-parametric prediction interval methods to quantify the data uncertainty and avoid the explicit parametric format. Meanwhile,  given the limited training state space, model uncertainty also exists. The DNN model may not learn the optimum policy function and may miss the unexplored state spaces,  potentially giving higher rewards. 
This means that the DRL model faces a trade-off between exploitation and exploration. Exploitation means utilizing the model's knowledge and choosing the best policy to maximize future rewards. Exploration involves selecting unexplored states to learn about potential high-reward state-action pairs. The challenge of effective
exploration is connected to model uncertainty. The higher model uncertainty means the model has not learned well in the given state and requires more exploration of that sample. For example,  the deep ensemble Q-network  \cite{chen2021randomized} is proposed to inject model uncertainty into Q learning for more efficient exploration sampling.  To reduce computational overhead, the Dropout Q-functions \cite{hiraoka2021dropout} method uses MC-dropout for model uncertainty quantification. 
The following work \cite{osband2018randomized} demonstrates that the previous ensemble and dropout methods may produce a poor approximation to the model uncertainty in cases where state density does not correlate with the true
uncertainty. To overcome the shortcoming, they suggest adding a random prior to the ensemble DQNs.


{\color{black}
In summary, UQ plays a critical role across various machine-learning problems. In out-of-distribution detection, we categorize uncertainty due to domain shift as model uncertainty rather than data uncertainty. BNN, ensemble, and distance-based methods are well-suited in these cases by capturing model uncertainty through weight distributions or sample embeddings. In active learning, identifying samples with high model uncertainty is essential for improving model training, while samples with high data uncertainty are generally less important for sample selection. Therefore, BNN and ensemble-based approaches play a larger role in active learning. In deep reinforcement learning,  data and model uncertainty are important since both help in efficient policy learning and exploration. Therefore, the combined approach should be used in this context. }






\section{Future direction}

This section identifies several future directions, including UQ for large language models, UQ for deep learning in scientific simulations, combining UQ and explainability, and UQ for DNN with structured outputs.  

\subsection{UQ for Large Language Models}
{\color{black}
In recent years, large language models (LLMs) \cite{achiam2023gpt} have revolutionized deep learning across diverse tasks such as text summarization, machine translation, complex problem-solving, and even creative writing.
However, it is found that an LLM sometimes generates 
over-confident outputs that read plausible but are factually incorrect, nonsensical, or unsupported by their training data, also called \emph{hallucinations}~\cite{chen2024inside,fadeeva2023lm}. Designing UQ methods for LLMs is essential for improving trustworthiness, especially in high-stakes applications.
However, several unique challenges exist. Unlike simpler DNN models, LLMs operate in a high-dimensional output space of a long sequence of tokens, making traditional measures like standard token entropy insufficient \cite{lin2023generating}. Furthermore, LLM-generated outputs may vary lexically (different token sequences) but convey the same semantic meaning, requiring UQ that can assess semantic similarity \cite{farquhar2024detecting}.


\begin{table*}[ht]
\footnotesize
\centering
\caption{\color{black} Categorization of Existing Methods for Uncertainty Quantification in LLMs}
\small
\begin{tabular}{|>{\centering\arraybackslash}p{3cm}|>{\centering\arraybackslash}p{4.3cm}|>{\centering\arraybackslash}p{5.5cm}|}
\hline
\textbf{Strategy} & \textbf{Black-box Methods} & \textbf{White-box Methods} \\
\hline
 Token probability-based UQ & \begin{tabular}[c]{@{}l@{}} \\ N/A \\ \end{tabular} & Reweight token entropy based on token importance~\cite{duan2024shifting}, Claim-conditioned token uncertainty for factual claim detection~\cite{fadeeva2024fact} \\
\hline
Self-knowledge-based UQ & Use self-evaluation prompting to elicit a confidence score from the model~\cite{chen2024quantifying} & Train a separate module using latent representations to predict uncertainty~\cite{azaria2023internal} \\
\hline
Sampling-based  UQ & Generate multiple outputs and measure response similarity~\cite{lin2023generating} & Analyze response covariance in latent space (e.g., eigenvalues)~\cite{chen2024inside} \\
\hline
\end{tabular}
\label{tab:uq_methods}
\end{table*}

Existing methods for UQ in LLMs can be categorized based on their underlying strategies, including token probability-based,  self-knowledge-based, and sampling-based. Within each strategy, specific methods can also be divided into black-box methods and white-box-based methods, according to whether a method requires access to model internal details \cite{geng2023survey}. The categorization is summarized in Table~\ref{tab:uq_methods}. 
The first category, i.e., token-probability-based UQ, is only applicable in white-box settings, as it requires access to token-logit level outputs. One approach focuses on reweighting token class entropy based on the importance of each token~\cite{duan2024shifting}. Another method, Claim Conditioned Probability (CCP), quantifies token-level uncertainty specifically for factual claims, filtering out noise from uncertainty about claim formulation~\cite{fadeeva2024fact}.
The second category is self-knowledge-based UQ. For black-box methods, self-evaluation is often used to prompt an LLM to produce a confidence score~\cite{chen2024quantifying}. In white-box scenarios, a separate module can be trained to predict uncertainty scores based on the LLM’s latent representations~\cite{azaria2023internal}. The third category is sampling-based UQ. In black-box methods, multiple samples are generated to assess response similarity~\cite{ lin2023generating}. For white-box approaches,  methods may analyze response covariance in latent embeddings (e.g., examining eigenvalues) to measure semantic consistency \cite{chen2024inside}. Additionally, conformal prediction has also been used to produce a prediction set of possible outputs that include the correct answer with a specified error rate~\cite{ye2024benchmarking}. Some semantic consistency-based methods are general for both white-box and black-box scenarios. For instance, \cite{farquhar2024detecting} proposes computing semantic entropy, which can be measured either using hidden embeddings (white-box) or by analyzing prompt outputs (black-box) from an LLM.

Future research on UQ for LLMs can be broadly grouped into three categories: UQ methodologies, UQ evaluation, and UQ-enabled agentic AI applications.

(1) Advancing UQ methodologies:
A central direction is to develop UQ methods that better reflect the semantic and generative structure of LLM outputs while offering clearer theoretical interpretations. 
Promising avenues include semantic-level uncertainty measures that aggregate multiple generations into meaning-equivalent clusters, hidden-state–based probes that extract single-pass uncertainty signals, and sampling-based estimators such as self-consistency–induced variance \cite{huang2024survey}.
A key open challenge is improving the calibration of base LLMs for long-form and multi-step generation, where token-level confidence estimates are poorly aligned with semantic correctness.

(2) Improving UQ evaluation and assessment:
Beyond standard accuracy and calibration metrics such as AUC or ECE, future work should establish evaluation protocols tailored to the structured and open-ended nature of LLM outputs \cite{xieempirical}. This includes span-level and answer-level calibration metrics that capture localized uncertainty within multi-sentence or multi-step responses, as well as decision-oriented metrics that evaluate abstention or self-verification behavior at fixed risk levels. Benchmarking hallucination detection remains crucial, and evaluation suites should explicitly disentangle the effects of sampling strategies from the quality of uncertainty signals \cite{herrera2025overview}.

(3) Leveraging UQ in agentic AI applications:
A third major direction is to integrate UQ into downstream LLM agent systems as a control signal rather than a diagnostic output. In LLM agents, uncertainty estimates can modulate decision-making processes such as tool calling, retrieval depth, human escalation, and risk-aware planning under ambiguity \cite{zhao2025uncertainty}. More broadly, UQ enables safer deployment in high-stakes domains by supporting abstention, clarification, or verification behaviors when estimated risk exceeds acceptable thresholds \cite{kirchhof2025position}. An important open problem is how to propagate and compose uncertainty across multi-component agentic pipelines involving retrieval, reasoning, and action.

\subsection{UQ for deep learning in scientific simulations}

Effective and efficient simulation of scientific phenomena, such as extreme weather events, climate change, and tsunamis, often require running physical models \cite{li2024generative}. Traditionally, these physical models are based on numerical Partial Differential Equations (PDEs), which are computationally intensive. In recent years, scientific machine (deep) learning has emerged as a new paradigm since data-driven techniques can learn complex patterns from vast amounts of data and make fast predictions with GPUs \cite{kochkov2024neural}. UQ for deep learning in scientific simulation is crucial in high-stake decision-making applications (e.g., disaster response). The uncertainty in scientific simulation and modeling can come from different sources. First, the initial and boundary conditions of the physical system are non-deterministic, and the system may be chaotic~\cite{wang2019deep}.  Second, the inherent physical principle may be imperfectly known, or the parameter of the governing equation may be stochastic. For example, the conservation law of heat may be violated in a non-closed system \cite{yang2019adversarial}.  Compared to traditional physics-based numerical simulations, diffusion models can generate ensembles of predictions more quickly for uncertainty estimation through probabilistic sampling \cite{li2024generative}. 

Physics-informed neural networks (PINNs) \cite{karniadakis2021physics} and neural operators \cite{kovachki2023neural} are currently two major deep learning techniques for solving PDEs in scientific simulations. PINNs incorporate physical constraints as soft regularization within the loss function, ensuring adherence to governing physical laws. In contrast, neural operators aim to train a neural network surrogate for a family of PDE instances. To enable uncertainty quantification, PINNs, and neural operators are often combined with UQ methods such as Bayesian neural networks and ensemble approaches \cite{sahin2024deep, psaros2023uncertainty}. However, most existing works often focus on synthetic data instead of complex real-world applications (e.g., physical oceanographers). In recent years, deep generative models, especially diffusion models \cite{ho2020denoising}, are increasingly used for real-world scientific simulations such as weather forecasting~\cite{price2023gencast, li2024generative, gao2024generative}. 

There are several potential future research directions. One direction is to decompose different sources of uncertainty, including those from model misspecification, stochasticity, incomplete knowledge of the underlying physical processes, and uncertainties tied to initial conditions, boundary conditions, and external forcings. Second, more efforts are needed for UQ for AI in simulating and forecasting extreme events, such as storm surges \cite{tebaldi2006going}. These events are rare (less observational data are available for training) but their societal impacts are very high. Moreover, model outputs can be highly sensitive to inputs (e.g., a small change in the input wind field and air pressures from a hurricane track will make a dramatic difference in output surge levels). Addressing this challenge requires the incorporation of physical knowledge in the UQ framework of the AI surrogate. Another direction is to improve the computational efficiency of AI models such as diffusion models, which are slow for both training and inference due to a large number of iterations \cite{ho2020denoising}. This is of particular importance for high-resolution spatiotemporal simulations. Finally, it is important to design UQ methods for AI in long-term temporal forecasting as error and uncertainty can accumulate over extended time horizons \cite{kochkov2024neural}. 
 
\subsection{Combine UQ with DNN explainability}

The explanation for DNN model predictions has been increasingly crucial because it provides tools for understanding and diagnosing the model's prediction. Recently, many explainability methods, termed explainers \cite{vu2020pgm}, have been introduced in the category of local feature attribution methods. That is, methods that return a real-valued score for each feature
of a given data sample, representing the feature’s relative importance for the sample prediction. These explanations are local in that each data sample may have a different explanation. Using local feature attribution methods, therefore, helps users better understand nonlinear and
complex black-box models. Both uncertainty quantification and explanation are important for a robust, trustworthy AI model. Current methodologies consider two directions separately, and we believe it could enable a more trustworthy AI system if combined.  Though many methodologies have been proposed for more precise uncertainty quantification, very few techniques attempt to explain why uncertainty exists in the predictions. 

{\color{black}
There are two possible directions to combine the power of explanations and uncertainty quantification: First, existing explanation methods could be potentially improved after considering the prediction uncertainty since those uncertain samples' explanations may not be trustworthy and can be omitted \cite{zhang2022explainable}. Second, we can leverage existing post hoc explanation methods to understand the source of the uncertainty \cite{jiang2024quantifying}. For example, it is intriguing to ask the question of why the prediction is uncertain and which set of input features are uncertain, or due to which layer of the model is imperfect. }

\subsection{UQ for DNNs with structured outputs}
Structured data are samples that are interdependent with each other, violating the common i.i.d assumption~\cite{bakir2007predicting}. Examples are imaging data, spatiotemporal data, and graphs. Deep learning has been widely used to model structured data, but the uncertainty of its prediction is not often quantified.
Here, we list future research directions for the three different types of structured data. 

\subsubsection{Imaging and inverse problem}

The goal of the imaging process is to reconstruct an unknown image from measurements, which is an inverse problem commonly used in medical imaging (e.g., magnetic resonance imaging and X-ray computed tomography) and geophysical inversion (e.g., seismic inversion) \cite{ edupuganti2020uncertainty}. However, this process is challenging due to the limited and noisy information used to determine the original image, leading to structured uncertainty and correlations between nearby pixels in the reconstructed image \cite{kendall2017uncertainties}. To overcome this issue, current research in uncertainty quantification of inverse problems employs conditional deep generative models, such as cVAE, cGAN, and conditional normalizing flow models \cite{dorta2018structured}. These methods utilize a low-dimensional latent space for image generation but may overlook unique data characteristics, such as structural constraints from domain physics in certain types of image data, such as remote sensing images, MRI images, or geological subsurface images \cite{jiang2017spatial, shih2022uncertainty}. The use of physics-informed models may improve uncertainty quantification in these cases. It's promising to incorporate the physics constraints for quantifying the uncertainty associated with the imaging process.

\subsubsection{Spatiotemporal data}
Spatiotemporal data are special due to the violation of the common assumption that samples follow an identical and independent distribution~ \cite{shekhar2015spatiotemporal,jiang2018survey}. Uncertainty quantification of spatiotemporal deep learning poses several unique challenges. First, the analysis of spatiotemporal data requires the co-registration of different maps (e.g., points, lines, polygons, geo-raster) into the same spatial reference system. The process is subject to registration uncertainty due to GPS errors or annotation mistakes in map generation~ \cite{jiang2021weakly}. Such registration uncertainty causes troubles when training deep neural networks~ \cite{he2022quantifying}. Second, implicit dependency structures exist in continuous space and time (e.g., spatial and temporal autocorrelation, and temporal dynamics). Thus, the uncertainty quantification process should be aware of such a dependency structure. Third, spatiotemporal non-stationary requires characterizing uncertainty due to out-of-distribution samples~\cite{shekhar2015spatiotemporal,jiang2018survey}. In addition, a different level of uncertainty exists based on the nearby training sample density. Traditionally, the Gaussian process has been widely used to quantify spatial uncertainty. However, for deep neural network models, new techniques are needed that consider sample density both in the non-spatial feature space and in the geographic space. 

\subsubsection{Graph data}
Graph data is a general type of structured data with nodes and edge connections. Graph neural networks (GNNs) have been widely used for graph applications related to node classification and edge (link) prediction. However, UQ for GNN models has been less explored. Some work utilizes existing UQ techniques for GNN models~ \cite{feng2021uag} without considering their unique characteristics. First, predictions on a graph are structured, so the UQ module needs to consider such structural dependency. Second, many GNN models assume a fixed graph topology from training and test instances (e.g., spectral-based methods~ \cite{defferrard2016convolutional}). Uncertainty in GNN predictions arises from shifts in graph topology between training and test graphs. Similarly, uncertainty exists when the graph is perturbed by removing nodes and edges. 
Finally, many real-world graph problems are spatiotemporal at the same time (e.g., traffic flow prediction on road networks). Thus, challenges related to UQ for spatiotemporal deep learning also apply to graphs. 


{\color{black}

\section{Conclusion}
This paper presents a systematic survey on uncertainty quantification for DNNs based on the types of uncertainty sources. We categorize the existing literature into three groups: model uncertainty, data uncertainty, and their combination. Additionally, we analyze the strengths and weaknesses of each approach based on the specific type of uncertainty it addresses. We also summarize the sources of uncertainty and the unique challenges faced across various applications, and ML problems, and propose several future research directions.
\section*{Acknowledgement}
This material is based upon work supported by the National Science Foundation (NSF) under Grant No. IIS-2147908, IIS-2207072, OAC-2152085, OAC-2402946, and OAC-2410884.



\bibliographystyle{ACM-Reference-Format}
\bibliography{ref}


\end{document}